\pdfoutput=1
\documentclass[11pt]{article}

\usepackage{EMNLP2022}

\usepackage{times}
\usepackage{latexsym}

\usepackage[T1]{fontenc}
\usepackage[utf8]{inputenc}

\usepackage{microtype}

\usepackage{inconsolata}

\usepackage{amsmath, amssymb}
\usepackage{bbm}
\usepackage{booktabs}
\usepackage{enumitem}
\usepackage{xspace}
\usepackage{graphicx}
\usepackage{caption}
\usepackage{subcaption}
\usepackage{listings}
\newcommand{\interalia}[1]{\citep[\textit{inter alia}]{#1}}

\newcommand{\mbrexec}{\textsc{MBR-exec}\xspace}
\newcommand{\mbrbleu}{\textsc{MBR-bleu}\xspace}
\newcommand{\maxlikelihood}{ML\xspace}
\newcommand{\maxavglikelihood}{\textsc{MaLL}\xspace}

\title{Natural Language to Code Translation with Execution}
\author{
Freda Shi$^{1, 2, }$\thanks{~~Work done while interning at Meta AI.} \quad
Daniel Fried$^{1, 3}$  \quad
Marjan Ghazvininejad$^{1}$
\\
\textbf{%
Luke Zettlemoyer$^{1, 4}$ \quad
Sida I. Wang$^{1}$}
\\
$^1$Meta AI \quad $^2$Toyota Technological Institute at Chicago \\
$^3$Carnegie Mellon University \quad $^4$University of Washington\\
\texttt{freda@ttic.edu} \quad \texttt{dfried@cs.cmu.edu} \quad \texttt{\{ghazvini,lsz,sida\}@fb.com}
}

\begin{document}
\maketitle
\definecolor{airforceblue}{rgb}{0.36, 0.54, 0.66}
\begin{abstract}
    Generative models of code, pretrained on large corpora of programs, have shown great success in translating natural language to code \interalia{chen2021evaluating,austin2021program,li2022competition}.
    While these models do not explicitly incorporate program semantics (i.e., execution results) during training, they are able to generate correct solutions for many problems. 
    However, choosing a \emph{single} correct program from a generated set for each problem remains challenging.

    In this work, we introduce execution result--based minimum Bayes risk decoding (\mbrexec) for program selection and show that it improves the few-shot performance of pretrained code models on natural-language-to-code tasks. We select output programs from a generated candidate set by marginalizing over program implementations that share the same semantics. Because exact equivalence is intractable, we execute each program on a small number of test inputs to approximate semantic equivalence. Across datasets, execution or simulated execution significantly outperforms the methods that do not involve program semantics. We find that \mbrexec consistently improves over all execution-unaware selection methods, suggesting it as an effective approach for natural language to code translation.\footnote{We open source our code at \href{https://github.com/facebookresearch/mbr-exec}{\textcolor{airforceblue}{[this URL]}} and our data at \href{https://dl.fbaipublicfiles.com/mbr-exec/mbr-exec-release.zip}{\textcolor{airforceblue}{[this URL]}}.}
\end{abstract}
\section{Introduction}
\label{sec:intro}
\begin{figure}[t]
    \centering
    \includegraphics[width=0.5\textwidth]{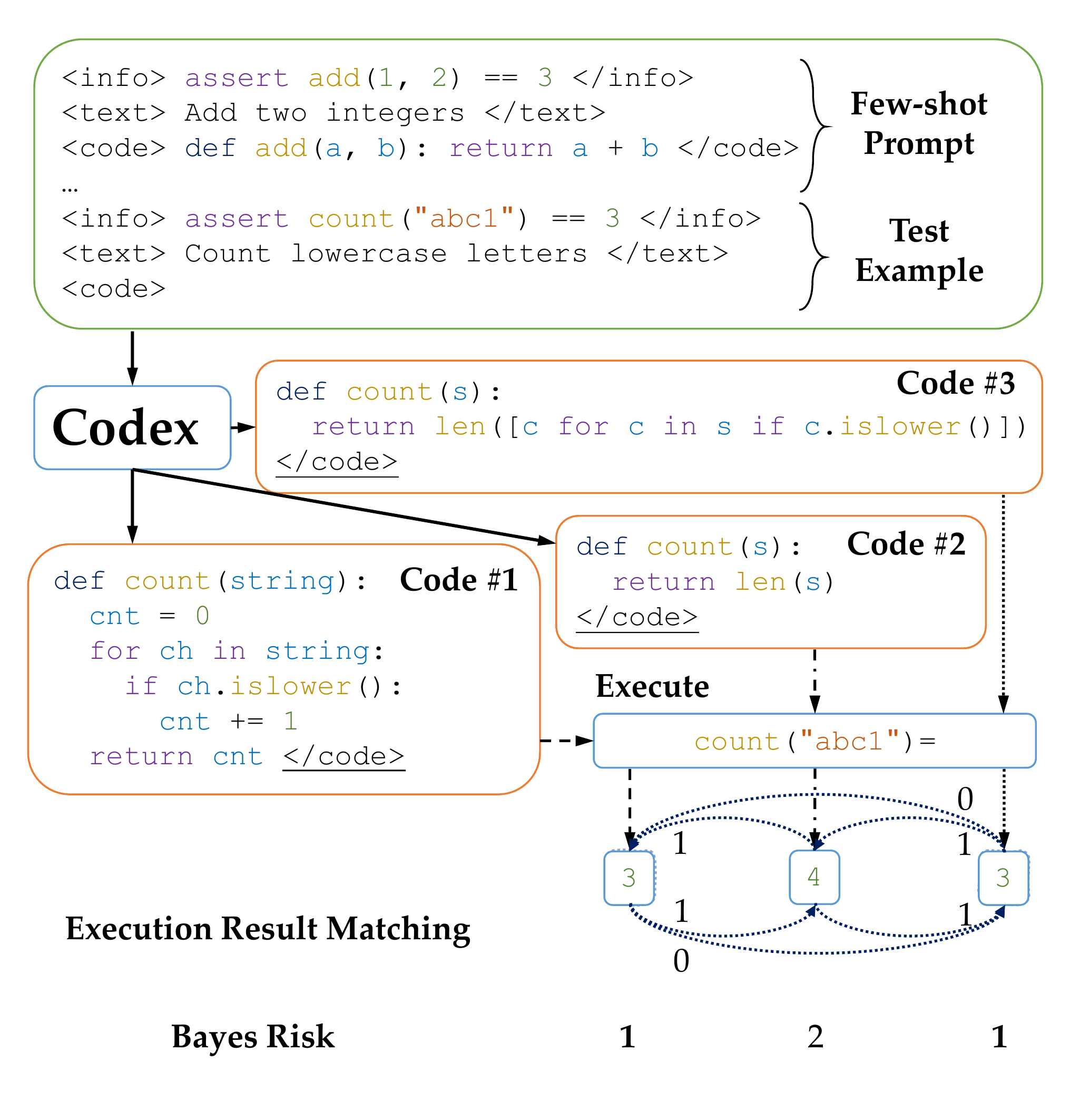}
    \caption{Illustration of \mbrexec on translating natural language to Python code: we (1) sample programs from Codex \citep{chen2021evaluating}, (2) execute each program on one test case, and (3) select the example with the minimal execution result--based Bayes risk. 
    Numbers around dotted lines denote the 0/1 matching loss between execution results, while the Bayes risk of a program is defined by the sum of the loss between itself and other examples. 
    In the figure, either Code \#1 or Code \#3 can be selected. Ground-truth program output is not needed for selection.} 
    \label{fig:teaser}
\end{figure}

The recent success of large pretrained language models \citep{radford2019language,brown2020language} has extended to translating natural language descriptions into executable code \interalia{chen2021evaluating,austin2021program,li2022competition}. After pretraining on large corpora of code with a simple language modeling objective,
%that maximizes the log likelihood of the next token conditioned on the existing ones,
the models demonstrate the ability to follow few-shot prompts \citep{radford2019language,brown2020language} to translate natural language to various programming languages. While code sampled from such models obtains surprisingly good BLEU scores against ground-truth programs and relatively high execution accuracies, it often includes obvious mistakes, and is of much lower quality than the code written by intermediate-level human programmers \citep{li2022competition}. 
In addition, choosing a \textit{single} correct one from a set of generated programs remains challenging.

In this work, we translate natural language to executable code with awareness of execution results on a limited number of test case inputs, which we require only at inference time. %while we only require gold test input when executing the programs.
% \dfried{recommend saying here that only test inputs are needed}\freda{Added. One thing to be in mind is that the prompt assertion for MBPP contains ground-truth output, though we didn't use it explicitly.}
Our approach is built on the hypothesis that a pretrained code model spreads probability mass over multiple semantically-equivalent code forms that implement the same functionality. Given a text description of a desired program function, we (1) sample a set of programs from a pretrained code model (\S\ref{sec:sample-collection}) and (2) select a single candidate program using execution-result-based minimum Bayes risk (MBR) decoding (\S\ref{sec:mbr-exec}). Intuitively, we score each sampled program using its agreement to other samples in terms of execution results, and select a program with maximal overall agreement. 

Our evaluation focuses on a challenging setting where only a single program can be submitted as the solution to a given problem. We show that the execution result--based selection method (i.e., \mbrexec) significantly outperforms all no-execution baselines across all considered datasets, despite having never executed any code during training and even when it has no access to ground-truth outputs. In addition, we show that MBR decoding with a BLEU-based risk function performs consistently well across datasets, and can be considered as a promising alternative when we are not able to execute. 

\begin{table*}[t]
    \centering \small
    \begin{tabular}{l|l}
        \toprule
        \textbf{\textit{General Template}} & \texttt{{<info>[INFO]</info>}} \qquad \textit{(optional)} \\
        & \texttt{{<text>[TEXT]</text>}} \\
        & \texttt{{<code>}[CODE]</code>} \\
        \midrule \midrule
        \textbf{\textit{Instantiation 1: Python}} & \texttt{<info>assert add(1, 2) == 3</info>}  \\
        \textbf{\textit{One-Shot Example}} & \texttt{<text>Write a function that adds 2 integers</text>} \\
        & \texttt{<code>def add(a, b): return a + b</code>} \\
        \midrule
        \textbf{\textit{Query}} & \texttt{<info>assert cat() == "cat"</info>}  \\
        & \texttt{<text>Write a function that outputs the string "cat"</text>} \\
        & \texttt{<code>} \\
        \midrule \midrule 
        \textbf{\textit{Instantiation 2: Bash}} & \texttt{<text>show the files in the current directory</text>} \\
        \textbf{\textit{One-Shot Example}} & \texttt{<code>ls</code>} \\
        \midrule
        \textbf{\textit{Query}} & \texttt{<text>show the first 5 lines of a.txt</text>}  \\
        & \texttt{<code>} \\
        \bottomrule
    \end{tabular}
    \caption{\label{tab:prompt-format} Prompt formatting template for queries to pretrained code models. For instantiation, we substitute \texttt{[TEXT]} and \texttt{[CODE]} with natural language descriptions and corresponding code snippets respectively. We also provide compatibility for an optional \texttt{[INFO]} section to provide the model extra information (e.g., the desired function identifier and example function calls) that helps code generation. In general, we expect the pretrained code models to generate a \texttt{</code>} token at the end of each code snippet given its pattern following ability \citep{brown2020language,chen2021evaluating}, otherwise we truncate the generated code to a maximum of 1024 tokens. }
\end{table*}

\section{Related Work}
\label{sec:related}

\subsection{Language to Code with Neural Networks}
\label{sec:related-lm-synthesis}
With the progress of neural network--based language modeling and conditioned text generation, there has been much work exploring natural language to code generation with end-to-end neural model architectures \interalia{xiao-etal-2016-sequence,ling-etal-2016-latent,rabinovich-etal-2017-abstract,dong-lapata-2018-coarse,suhr-etal-2018-learning,xu-etal-2020-incorporating,lachaux2021dobf}. 
Recently, large Transformer-based~\citep{vaswani2017attention} pretrained code models have shown surprisingly strong generation performance across programming languages \interalia{chen2021evaluating,austin2021program,li2022competition}. In this work, we explore selection (i.e., inference) methods to apply to these pretrained models, showing that selecting programs using their execution results can greatly improve program generation. 

Multiple benchmarks have been proposed to evaluate code model performance \interalia{miceli-barone-sennrich-2017-parallel,yin2018learning,hendrycks2021measuring,lu2021codexglue}. In this work, we evaluate on three text-to-code datasets: MBPP \citep[Python;][]{austin2021program}, Spider \citep[SQL;][]{yu-etal-2018-spider} and NL2Bash \citep[Bash;][]{lin-etal-2018-nl2bash}, covering a range of programming languages. 

\subsection{Prompting Pretrained Language Models}
The GPT-2 \citep{radford2019language} and GPT-3 \citep{brown2020language} models have shown strong prompting performance: after conditioning on a task-related prompt, the language models are often able to make accurate output predictions for unseen inputs. These results lead to prompt-based approaches for few-shot or zero-shot text classification \interalia{shin-etal-2020-autoprompt,gao-etal-2021-making,min2021noisy}, question answering \citep{khashabi-etal-2020-unifiedqa}, machine translation \citep{radford2019language}, and evaluation of generated text \citep{yuan2021bartscore}, where no more than a few examples are used to construct the prompts. Few-shot examples are usually formatted into natural language prompts and continuations generated by the models for these prompts are then converted to task-specific predictions. The prompt formatting can be either manually designed \citep{jiang-etal-2020-know} or automatically learned \citep{li-liang-2021-prefix,lester-etal-2021-power}. Recently, \citet{wang2022self} find that self-consistency based decoding improves chain-of-thought prompting \citep{wei2022chain}.
We refer the readers to \citet{liu2021pre} for a more comprehensive survey. 

In this work, we prompt a pretrained code model \citep[Codex;][]{chen2021evaluating} in a few-shot setting (\S\ref{sec:sample-collection}) and perform execution-based selection over the samples. We also find that the Codex model performs well with a fairly programming-language-agnostic prompt formatting (Table~\ref{tab:prompt-format}). 

\subsection{Minimum Bayes Risk Decoding}
In structured prediction, Minimum Bayes risk (MBR) decoding \citep{bickel1977mathematical} selects a structured output that minimizes the expected errors in the structure by introducing an explicit loss function to the decision criterion. This method has outperformed the maximum a posteriori (MAP) method on many tasks, including syntactic parsing \citep{titov2006bayes,shi-etal-2019-visually,zhang-etal-2020-efficient}, statistical machine translation \citep{kumar-byrne-2004-minimum,zhang-gildea-2008-efficient}, and neural machine translation \citep{eikema-aziz-2020-map,eikema2021sampling}. 

In machine translation, MBR decoding is usually implemented by reranking candidates \interalia{goel-byrne-2000-minimum, kumar-byrne-2004-minimum, tromble-etal-2008-lattice}. Let $F$ denote the input, and $E$ denote the corresponding ground-truth translation. Given a loss function $\ell(\cdot, \cdot)$ between translations and a probability model $P(E\mid F)$, MBR decoding can be formulated as 
\begin{align}
    \hat{E} = \arg\min_{E'\in \mathcal{E}_h} \sum_{E\in\mathcal{E}_e} \ell(E, E') P(E\mid F),
    \label{eq:mbr-decoding}
\end{align}
where $\mathcal{E}_h$ is the \textit{hypothesis space}, and $\mathcal{E}_e$ is the \textit{evidence space}: both are sets of possible translations. 

We define execution based MBR loss functions, and show that they are crucial in the sample selection processes for natural language to code with a pretrained large language model.

\section{Proposed Approach: \mbrexec}
Our execution-based framework consists of two parts: (1) collecting samples from a pretrained code model (\S\ref{sec:sample-collection}) and (2) selecting the best candidate using minimum Bayes risk decoding (\S\ref{sec:mbr-exec}).
\subsection{Sample Collection}
\label{sec:sample-collection}
To obtain the corresponding code, we query the pretrained code model with few-shot prompts followed by the text description, using a unified mark-up style few-shot prompting template (Table~\ref{tab:prompt-format}).\footnote{While existing work on prompting language models usually requires a task-specific design of prompts \interalia{shin-etal-2020-autoprompt,zhong-etal-2021-factual,gao-etal-2021-making}, we find that a fairly general pattern (Table~\ref{tab:prompt-format}), which does not involve any programming language--specific information, works well across programming languages on Codex.} 
In addition to the generated programs themselves, most existing models also allow us to have the associated probability of generating each generated token $w_i$ conditioned on the prompt tokens $C=\langle c_1, \ldots, c_n \rangle$ and all the previously generated tokens $w_1, \ldots, w_{i-1}$, denoted by $P(w_i \mid C, w_1, \ldots w_{i-1})$. 

\subsection{Execution-Based MBR Decoding}
\label{sec:mbr-exec}
Given a problem in its natural language description $C$, we sample a set of programs $\mathcal{P}=\{p_i\}_{i=1}^N$ using the method in \S\ref{sec:sample-collection}. We formulate the execution-based MBR (\mbrexec) decoding by selecting
\begin{align}
\hat{p} &= \arg\min_{p \in \mathcal{P}} \mathcal{L}_\textit{MBR}(p; \mathcal{P}) \nonumber \\
&=\arg\min_{p\in \mathcal{P}} \sum_{p_\textit{ref} \in \mathcal{P}} \ell(p, p_\textit{ref}) \label{eq:mbr-exec-decoding}
\end{align}
as the best candidate, where $\mathcal{L}_\textit{MBR}(\cdot; \cdot)$ denotes the MBR loss of a program conditioned on a set of references and $\ell$ is a predefined, execution-based loss function that examines the discrepancy between two programs. Intuitively, this finds a consensus candidate which has a low loss relative to all other candidates. The above implementation is an unbiased estimation of Eq~\eqref{eq:mbr-decoding}.

We introduce the following execution result--based loss function:
\begin{align*}
    \ell (p_i, p_j) &= \max_{t \in \mathcal{T}} \mathbbm{1}\left[p_i(t) \neq p_j(t)\right],
\end{align*}
where $\mathcal{T}$ is the set of available test inputs,\footnote{Our \mbrexec decoding process does not involve any ground-truth test case output, nor the ground-truth programs. This is compatible with many real scenarios, e.g., in a programming competition, where valid test input are easier to access than ground-truth output.} and $p_i(t)$ denotes the execution result of program $p_i$ when having $t$ as the input. When a program fails to execute on a test case, it is considered not equivalent to any other programs, even if they fail to execute as well. Intuitively, $\ell$ assigns equivalence ($0$ loss) if and only if two programs have the same output on all considered test cases.

There may be multiple programs receiving the same MBR loss $\mathcal{L}_\textit{MBR}(\cdot; \mathcal{P})$, which are all minima. We break any ties by selecting the program with the largest likelihood among them.

\section{Experiments}
\label{sec:expr}

\begin{figure*}[t!]
    \centering
    \begin{subfigure}[t]{0.34\textwidth}
        \includegraphics[width=\textwidth]{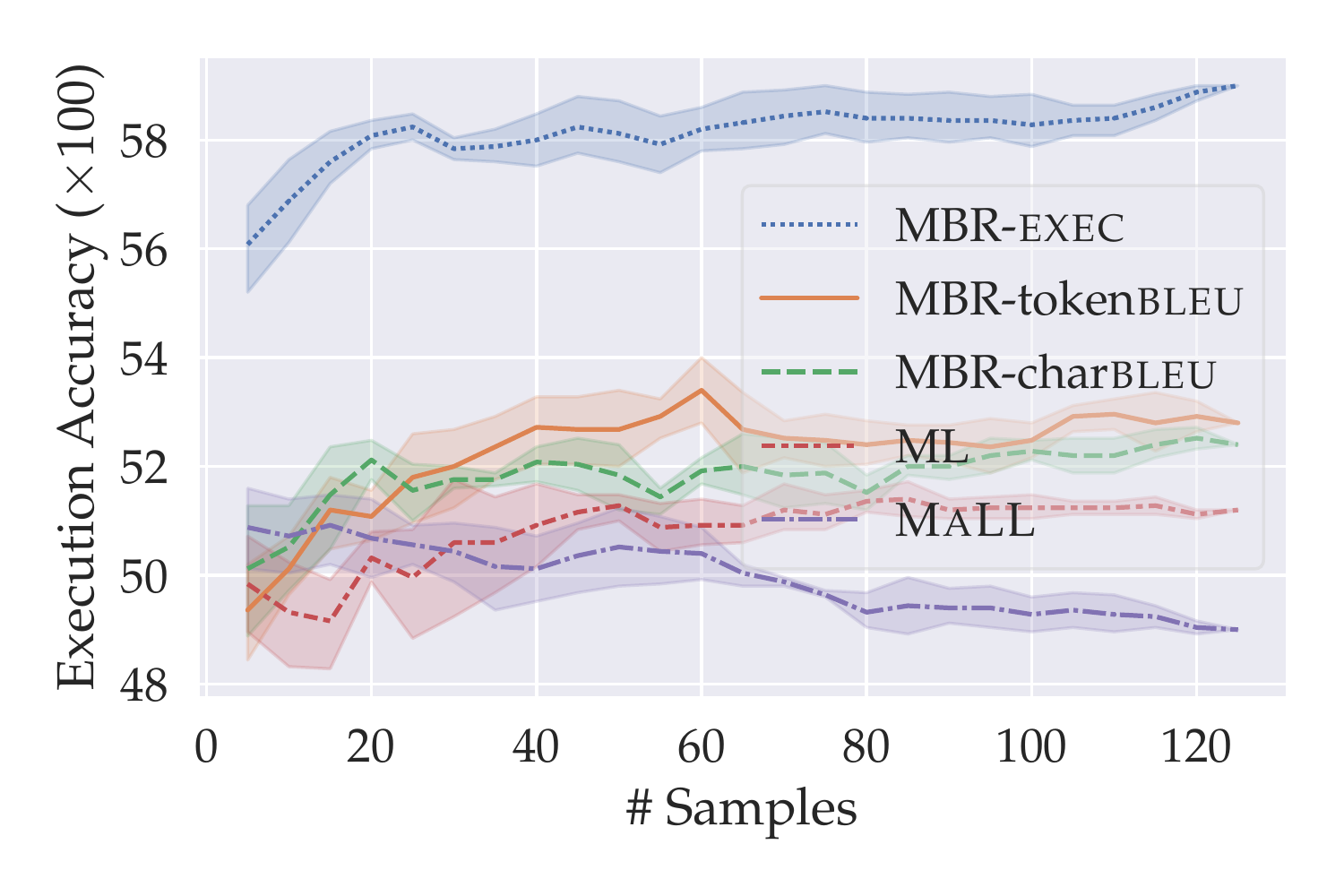}
        \caption{MBPP}
    \end{subfigure}
    \hspace{-10pt}
    \begin{subfigure}[t]{0.34\textwidth}
        \includegraphics[width=\textwidth]{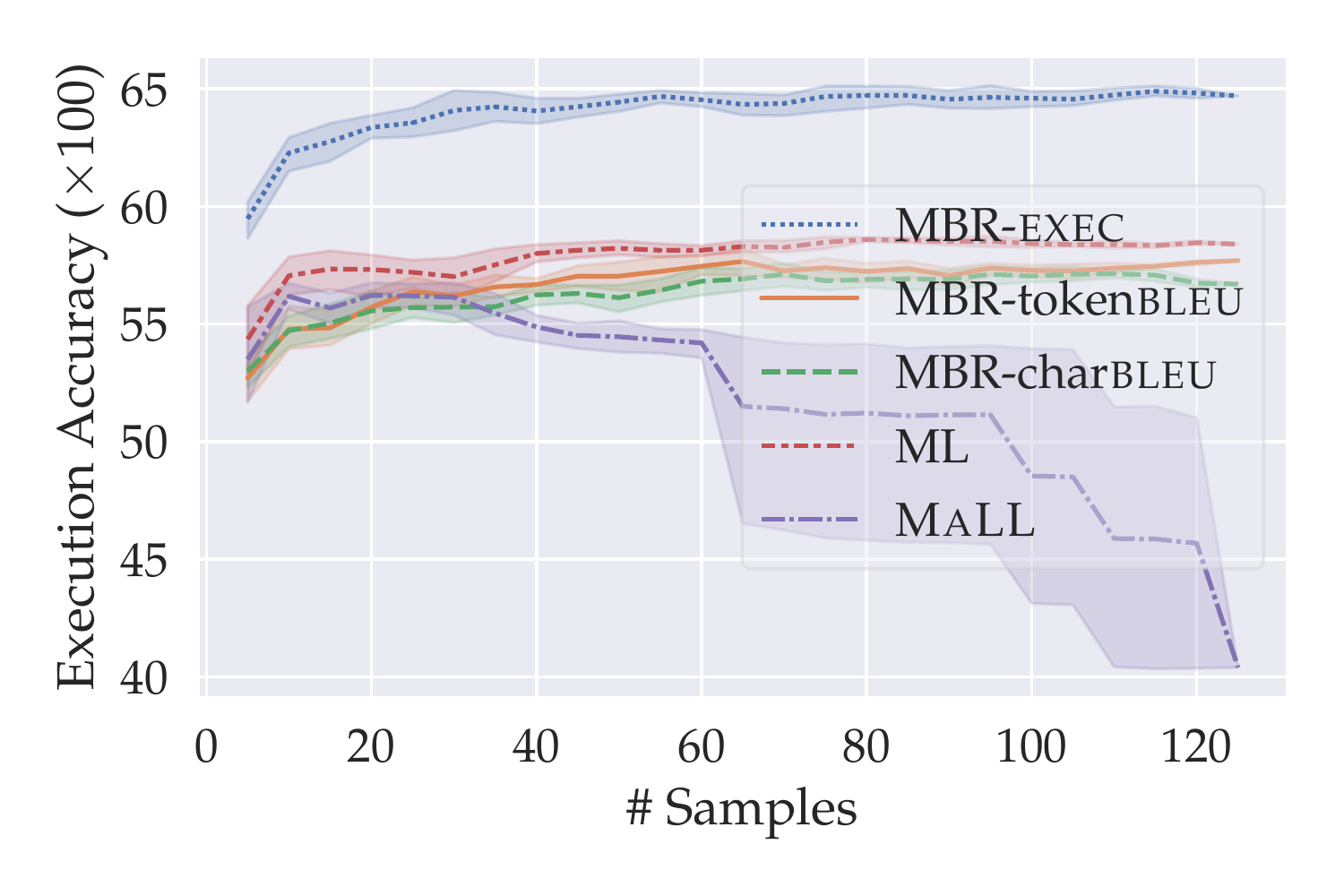}
        \caption{Spider}
    \end{subfigure}
    \hspace{-10pt}
    \begin{subfigure}[t]{0.34\textwidth}
        \includegraphics[width=\textwidth]{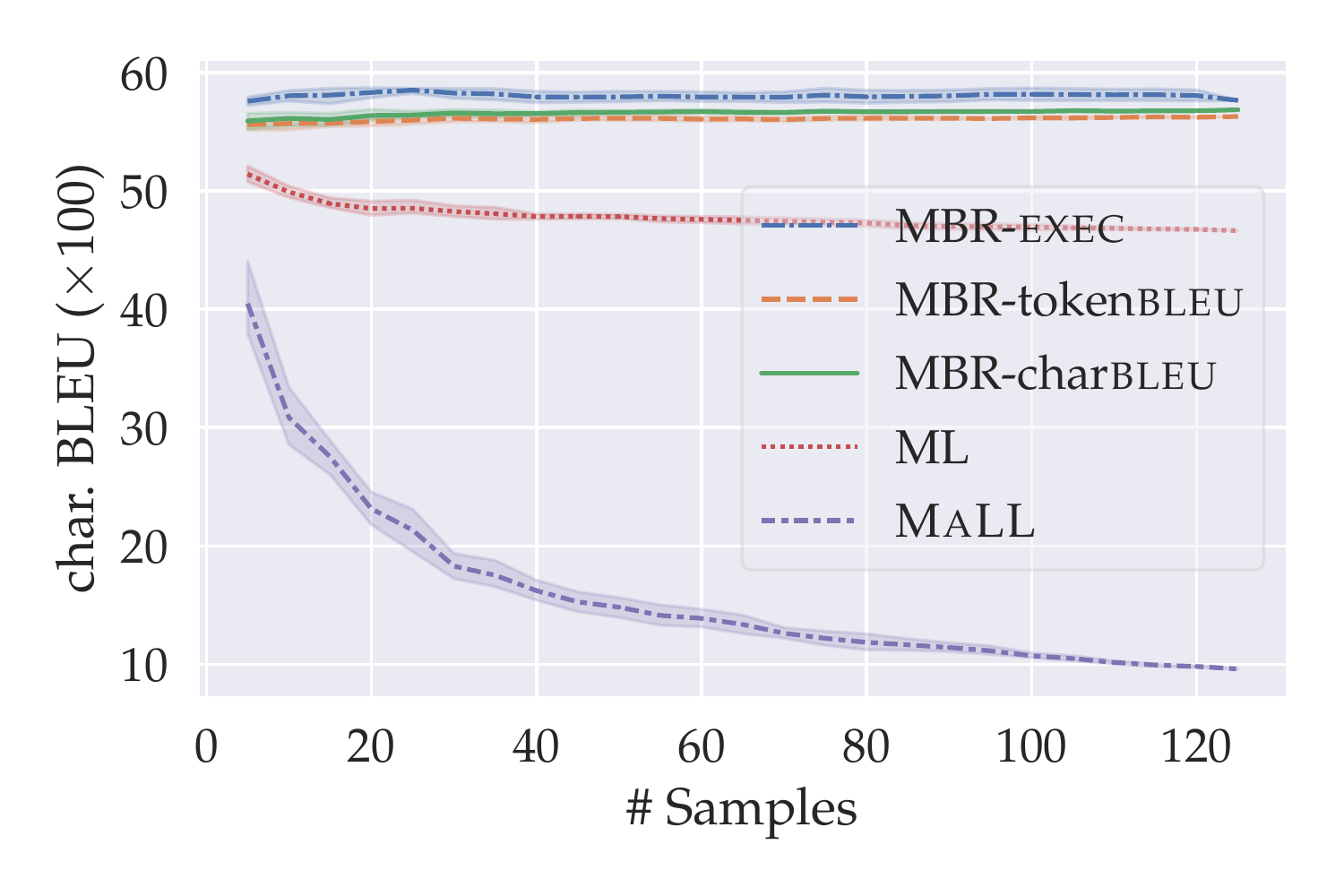}
        \caption{NL2Bash}
    \end{subfigure}
    \caption{\textbf{Primary evaluation results:} performance of the evaluated selection criteria (best viewed in color). For each sample size, we evaluate the methods on 5 different groups of samples and report the average performance (lines) and the standard deviations (shaded regions). All samples are collected from Codex with temperature 0.3. }
    \label{fig:main-results}
\end{figure*}

\begin{table}[t]
    \centering \small
    \begin{tabular}{lccc}
        \toprule
        \textbf{Method} & \textbf{MBPP} & \textbf{Spider} & \textbf{NL2Bash} \\
        \midrule
        Greedy (3-shot) &  $47.3\pm2.5$ \hspace{-6pt} & $50.8\pm2.6$ \hspace{-6pt}   & $52.8\pm2.9$ \\
        Sample (3-shot) &  $47.7\pm1.5$ \hspace{-6pt} & $48.5\pm2.6$ \hspace{-6pt}  &  $53.0\pm2.9$ \\
        \midrule 
        \mbrexec &  $\textbf{58.2}\pm0.3$ \hspace{-6pt} &  $\textbf{63.6}\pm 0.8$  \hspace{-6pt} &  $\textbf{58.5}\pm 0.3$ \\
        \bottomrule
    \end{tabular}
    \caption{Comparison between \mbrexec and baselines without selection process. For both \mbrexec and Sample (3-shot), we collected samples with temperature 0.3. All numbers involve the same set of 125 samples for each case: for greedy and sample baselines, we report average performance of them all; for \mbrexec, we report the result with 25 examples, averaged across 5 experiments. }
    \label{tab:main-results-basic-baseline}
\end{table}

We evaluate (\S\ref{sec:expr-results-main}) and analyze (\S\ref{sec:expr-analysis}) the performance of \mbrexec, starting with introducing the datasets and evaluation metrics (\S\ref{sec:expr-datasets}), as well as non-execution-based baselines (\S\ref{sec:expr-baselines}) for \mbrexec. Finally, we show and discuss oracle performances on the considered tasks (\S\ref{sec:expr-oracle}).
\subsection{Datasets and Evaluation Metrics}
\label{sec:expr-datasets}
We consider three datasets that cover a range of programming languages: MBPP \citep[Python;][]{austin2021program}, Spider \citep[SQL;][]{yu-etal-2018-spider}, and NL2Bash \citep[Bash;][]{lin-etal-2018-nl2bash}. 

\paragraph{MBPP.} 
The MBPP dataset \citep{austin2021program}\footnote{
    \url{https://github.com/google-research/google-research/tree/master/mbpp}
} 
consists of 974 basic Python programming problems, with 500 of them used for testing and the rest for training or few-shot prompting. There are ground-truth program and three assertions (i.e., test cases with input and ground-truth output) associated with the description of each problem. When collecting the samples, we use one assertion as the extra information (\texttt{[INFO]}; Table~\ref{tab:prompt-format}).\footnote{The main goal of \texttt{[INFO]} in MBPP is to inform Codex about the desired function name for easier evaluation -- while the assertions are not a necessary part of prompt, we use them as \texttt{[INFO]} for simplicity and compatibility with past work \citep{austin2021program}. } Programs are evaluated with execution accuracy, where a program is considered as passing if all three test cases are correct.

\paragraph{Spider.}
The Spider dataset \citep{yu-etal-2018-spider}\footnote{\url{https://yale-lily.github.io/spider}} is a text-to-SQL dataset, which requires a model to translate text descriptions into SQL commands. There are 7,000 examples for training and 1,034 for development. When prompting models to produce candidate commands, we concatenate the corresponding SQL table and column names as the \texttt{[INFO]}. Commands are evaluated with the execution accuracy, where a command is considered as passing if it returns the same result as the ground-truth command when being executed on the same database.

\paragraph{NL2Bash.} 
The NL2Bash dataset \citep{lin-etal-2018-nl2bash} aims to translate natural language to bash commands. We do not include \texttt{[INFO]} in the sample collection process. Because it is difficult to execute bash commands in a sandbox, we split a bash command with \texttt{bashlex},\footnote{\url{https://pypi.org/project/bashlex/}} a rule-based bash parser, and use the token-level BLEU-4 score between commands as the estimation of execution result similarity. We consider a command to be unexecutable when \texttt{bashlex} fails to parse it. Following \citet{lin-etal-2018-nl2bash}, commands are evaluated with character-level BLEU-4 score.

Across datasets, we use 15 examples from the training set for few-shot prompting. A detailed example showing prompt formatting can be found in Appendix~\ref{appendix:prompts}. Unless otherwise specified, we collect samples by querying Codex with five different prompts, each containing 3 examples, using temperature 0.3. We combine the candidates sampled across the five prompts to get a set of candidate samples to use in our selection methods. For execution on MBPP and Spider, we apply a memory limit of 128GB and a time limit of 10 seconds on a single Intel(R) Xeon(R) CPU E5-2698 v4 @ 2.20GHz CPU, and consider the programs that exceed these limits as inexecutable; unless otherwise specified, we only execute each program on the first test input provided for the example, and use the output for calculating the Bayes risk in the inference process. 

\subsection{Baselines}
\label{sec:expr-baselines}
We compare the most basic baselines with no selection, prompting Codex with three examples in Table~\ref{tab:prompt-format} format:\footnote{We use the \texttt{code-davinci-001} engine throughout this work.}
\begin{itemize}[leftmargin=*]\setlength{\itemsep}{0pt}
    \item \textbf{Greedy decoding.} We perform token by token greedy decoding to generate the output. 
    \item \textbf{Sampling.} We sample the output token by token with a fixed temperature, where we set the temperature as 0.3 in all of our experiments. 
\end{itemize}
In addition, we consider the following baseline sample selection methods:
\begin{itemize}[leftmargin=*]\setlength{\itemsep}{0pt}
    \item \textbf{Maximizing likelihood} (\maxlikelihood). Given a set of sampled candidate programs, we select the one with the largest log likelihood. Formally, we select 
    \begin{align*}
        \hat{p} = \arg\max_{p \in \mathcal{P}} \prod_{i=1}^{n_p} P(w_{p, i} \mid C, w_{p, 1}, \ldots, w_{p, i-1}),
    \end{align*}
    where $n_p$ denotes the number of tokens in a generated program $p$, and $w_{p, i}$ denotes its $i$-th token.
    \item \textbf{Maximizing average log likelihood} (\maxavglikelihood) across tokens. In order to address the practical issue that ML typically favors shorter sequences, we follow \citet{chen2021evaluating} and propose another baseline that uses the average log likelihood across tokens as the selection criterion, where we select 
    \begin{align*}
        \hat{p} = & \arg\max_{p \in \mathcal{P}} \\
        & \frac{1}{n_p} \sum_{i=1}^{n_p} \log P(w_{p, i} \mid C, w_{p, 1}, \ldots, w_{p, i-1}).
    \end{align*}
    \item \textbf{BLEU score based MBR} (\mbrbleu). To study the effect of execution based MBR in sample selection, we consider BLEU score based MBR, where the Bayes risk is calculated using the following risk function:
    \begin{align*}
        \ell_\textit{BLEU}(p_i, p_j) = -\text{BLEU}(p_i, p_j),
    \end{align*}
    where $\text{BLEU}(p_i, p_j)$ is the BLEU score of the two programs. We use character-level (MBR-char\textsc{bleu}) or token-level (MBR-token\textsc{bleu}) BLEU-4 in all of our experiments.
\end{itemize}

\subsection{Primary Results}
\label{sec:expr-results-main}
We evaluate \mbrexec on the three datasets (\S\ref{sec:expr-datasets}) with dataset-specific metric, where we use one test case for each problem. 
\mbrexec outperforms all baselines without a selection process by a significant margin (Table~\ref{tab:main-results-basic-baseline}).
In addition, we find that \mbrexec outperforms all baseline selection methods (Figure~\ref{fig:main-results}), and is especially effective on the two datasets (MBPP and Spider) that use execution-based evaluation. In addition, the \mbrbleu metrics are also strong and robust across datasets, suggesting the effectiveness of finding a consensus candidate that has generally low discrepancy with other samples. 

While more samples lead to better performance for most methods, \maxavglikelihood consistently performs worse with a larger sample size, as we find that \maxavglikelihood generally favors programs with unnecessary repetitions,\footnote{This issue has been found in existing open-ended text generation models, while methods such as unlikelihood training \citep{Welleck2020Neural} may help reduce degeneration (i.e., the generation of unnecessarily repetitive output).} and a larger sample size generally leads to a larger chance to have such a sample.

\begin{figure*}[t]
    \centering
    \begin{subfigure}[t]{0.34\textwidth}
        \includegraphics[width=\textwidth]{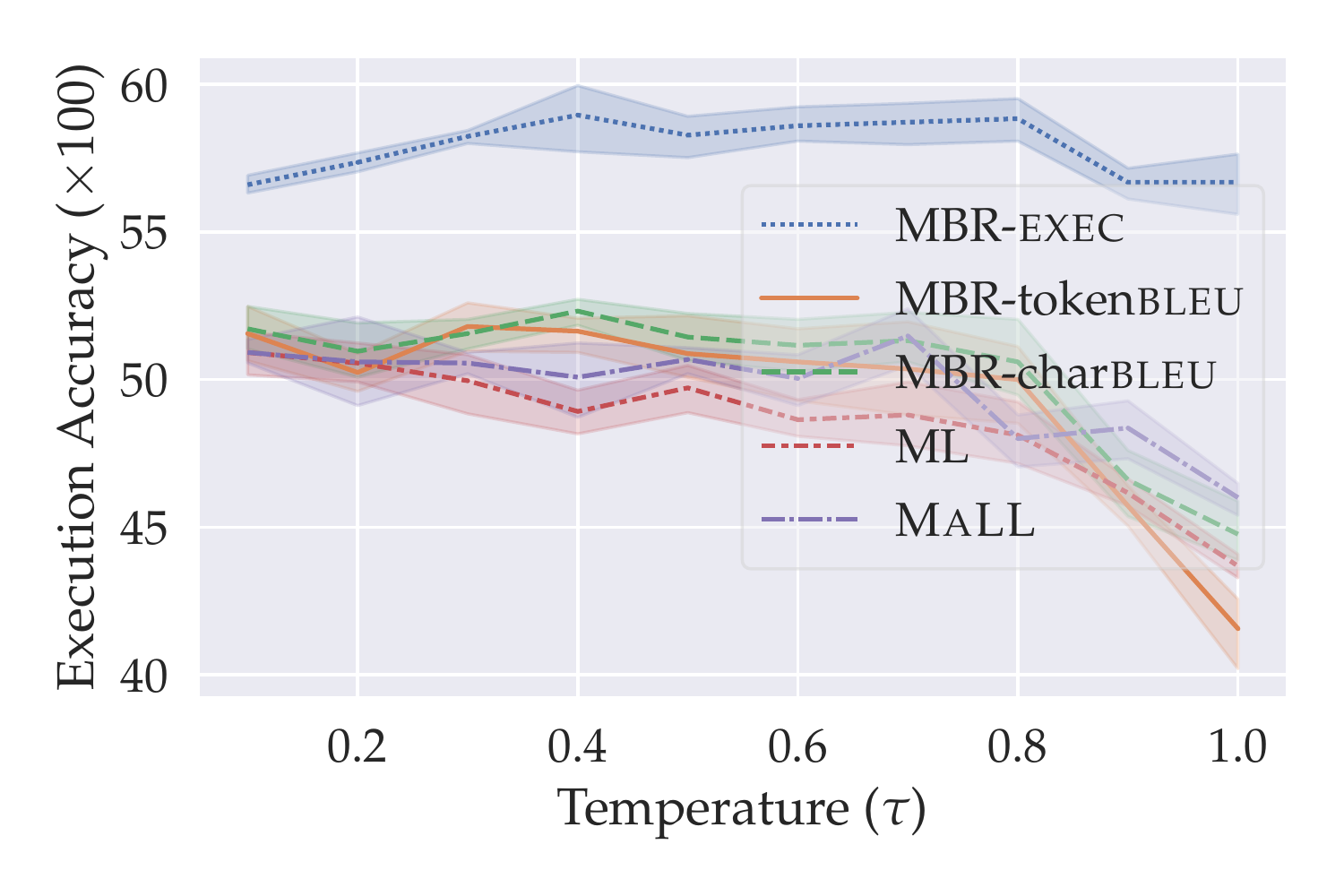}
        \caption{MBPP}
    \end{subfigure}
    \hspace{-10pt}
    \begin{subfigure}[t]{0.34\textwidth}
        \includegraphics[width=\textwidth]{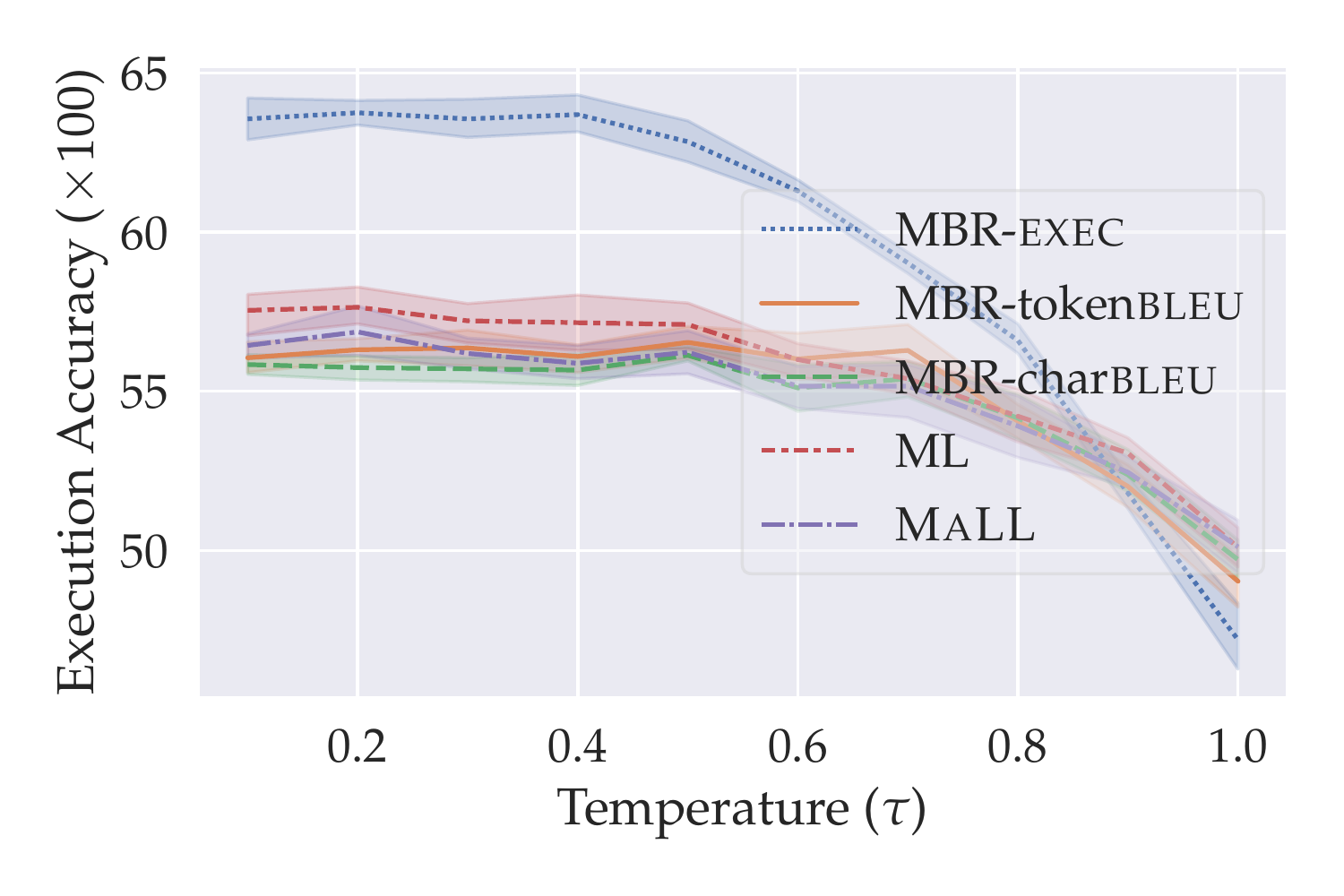}
        \caption{Spider}
    \end{subfigure}
    \hspace{-10pt}
    \begin{subfigure}[t]{0.34\textwidth}
        \includegraphics[width=\textwidth]{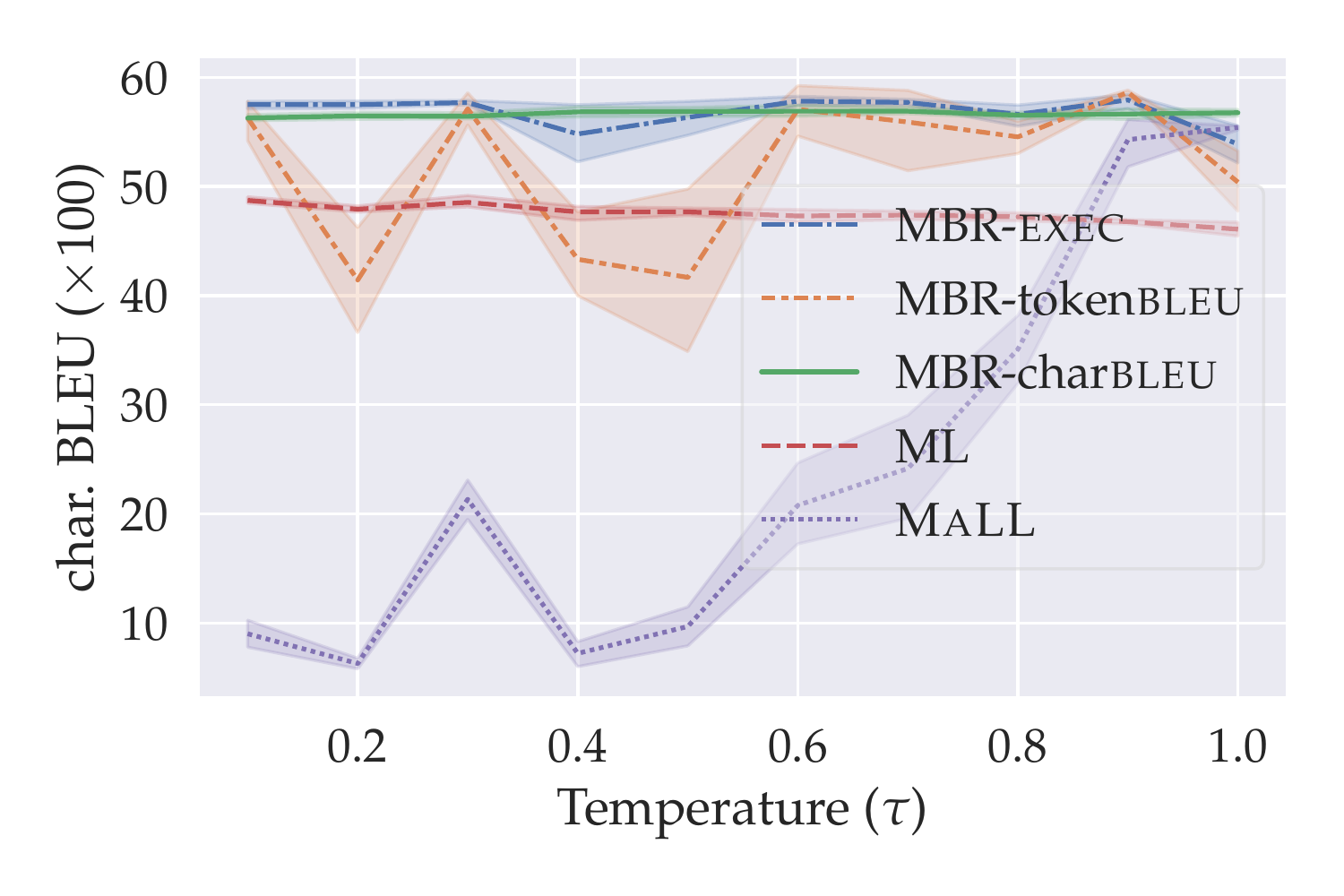}
        \caption{NL2Bash}
    \end{subfigure}
    \caption{Performance of the evaluated selection criteria across temperatures (best viewed in color). For each temperature, we perform the methods on 5 different groups of 25 examples and report the average performance (lines) and the standard deviations (shaded regions).}
    \label{fig:temperature}
\end{figure*}
\begin{table}[t]
\centering
\small
\begin{tabular}{lcc}
    \toprule
    \textbf{Dataset} & \textbf{Greedy ($\tau=0$)} & \textbf{Sample ($\tau=0.3$)} \\
    \midrule
    MBPP    & $56.0$    &  $\textbf{58.2}\pm0.3$\\
    Spider  & $62.1$    &  $\textbf{63.6}\pm0.8$\\
    NL2Bash & $58.4$    &  $\textbf{58.5}\pm0.3$\\
    \bottomrule
\end{tabular}
\caption{
    \label{tab:0.3-vs-greedy} \mbrexec performance on greedily decoded and sampled programs: for each problem, we use 25 groups of 3-shot prompts, decode or sample one program with each prompt, and use \mbrexec to select the best program.
    For sampling with temperature 0.3, we repeat the process for 5 times and report the average performance and standard deviations.
    The dataset-specific metric can be found at \S\ref{sec:expr-datasets}.
    The best number in each row is in boldface. 
    Note that the greedy performances are different from those reported in Table~\ref{tab:main-results-basic-baseline}, as we perform \mbrexec here over greedy decoding outputs, while report the average performance in Table~\ref{tab:main-results-basic-baseline}.
}
\end{table}
\begin{figure}[t!]
    \centering
    \begin{subfigure}[t]{0.48\textwidth}
        \includegraphics[width=\textwidth]{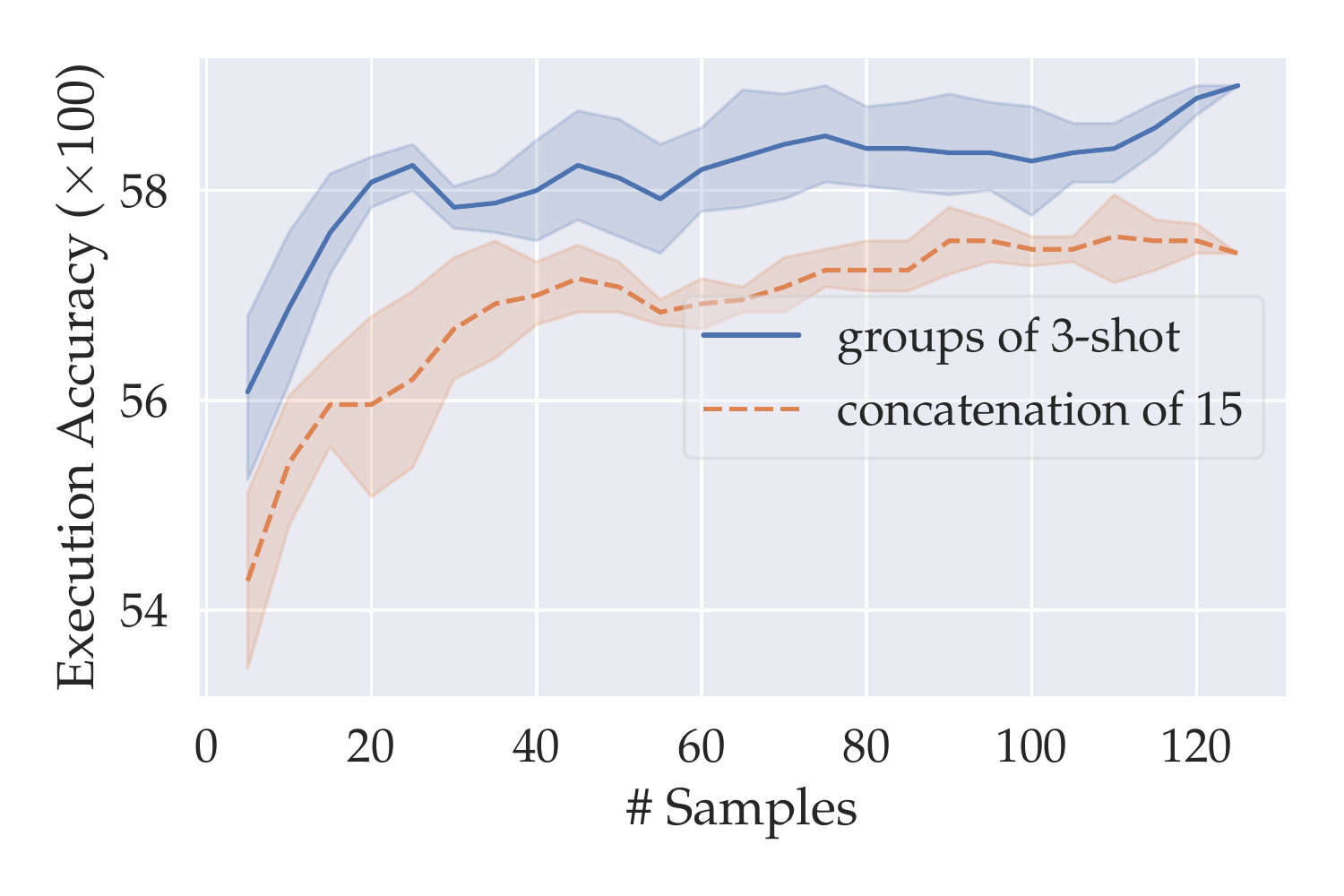}
        \caption{MBPP}
    \end{subfigure}
    \begin{subfigure}[t]{0.48\textwidth}
        \includegraphics[width=\textwidth]{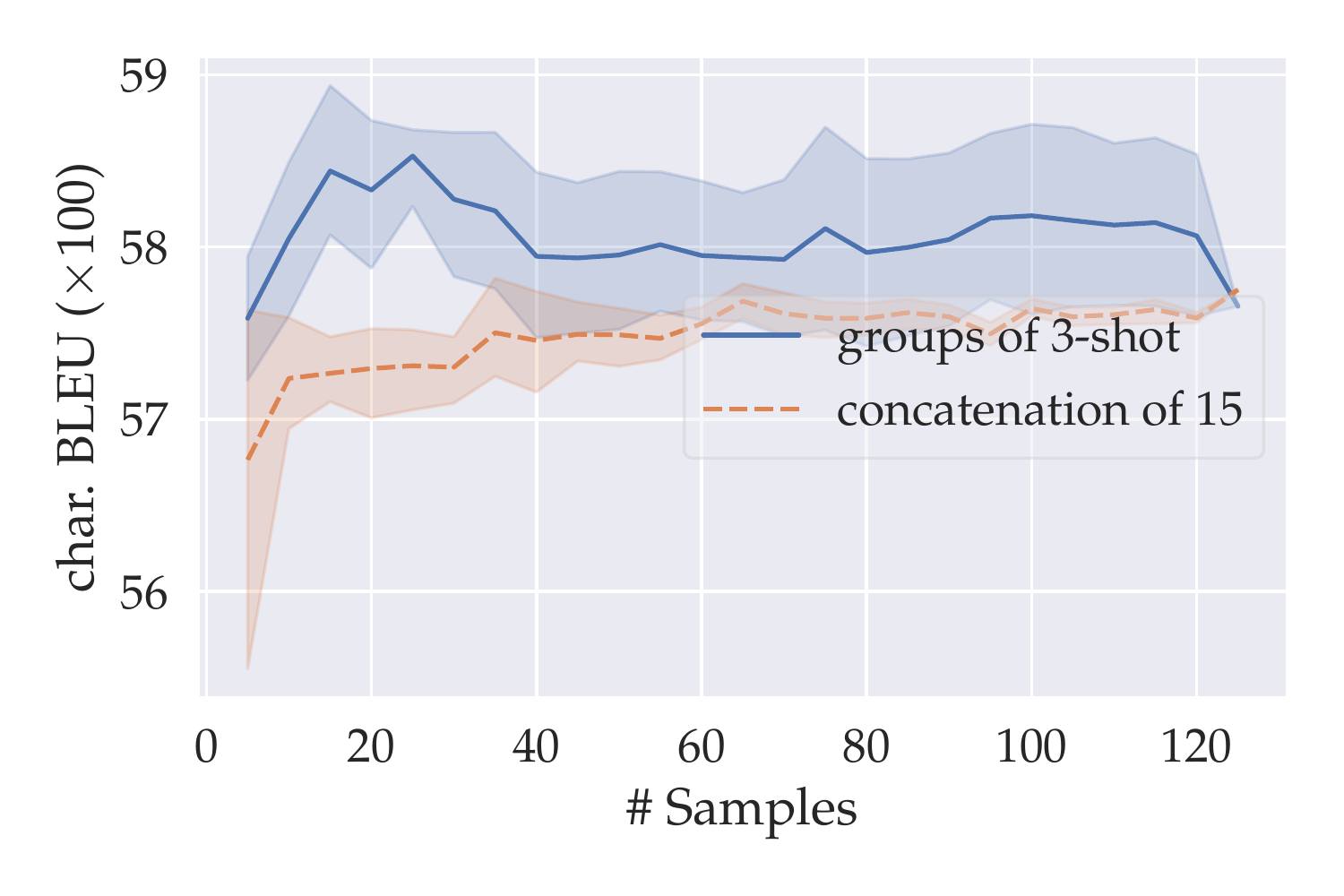}
        \caption{NL2Bash}
    \end{subfigure}
    \caption{Performance with different types of prompts, where \textit{groups of 3-shot} denotes the prompt formatting in Table~\ref{tab:prompt-format}, while \textit{concatenation of 15} denotes concatenating all available 15 examples as prompts for data collection. }
    \label{fig:15-shot}
\end{figure}
\begin{figure}[t]
    \centering
    \begin{subfigure}[t]{0.48\textwidth}
        \includegraphics[width=\textwidth]{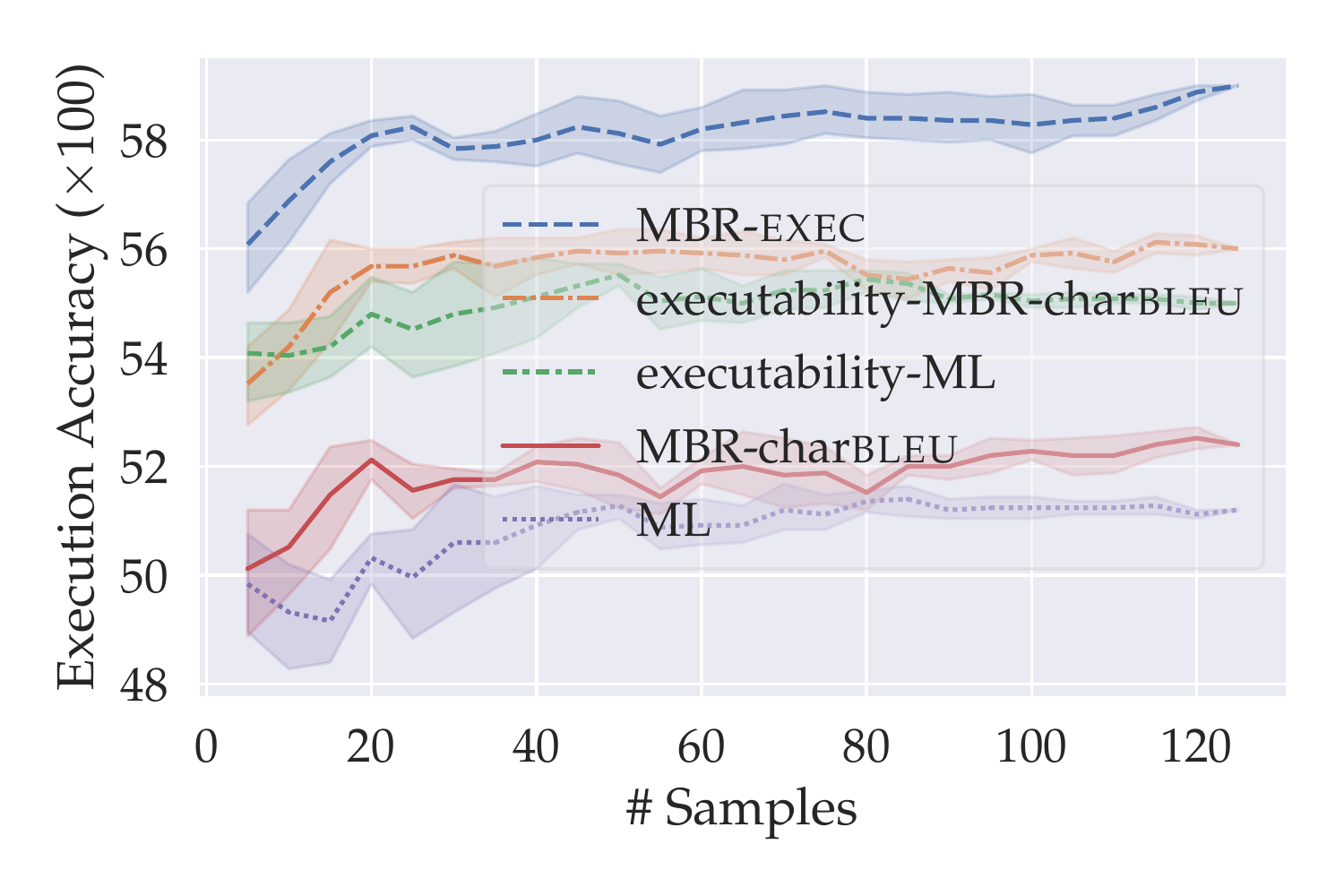}
        \caption{MBPP}
    \end{subfigure}
    \hspace{-10pt}
    \begin{subfigure}[t]{0.48\textwidth}
        \includegraphics[width=\textwidth]{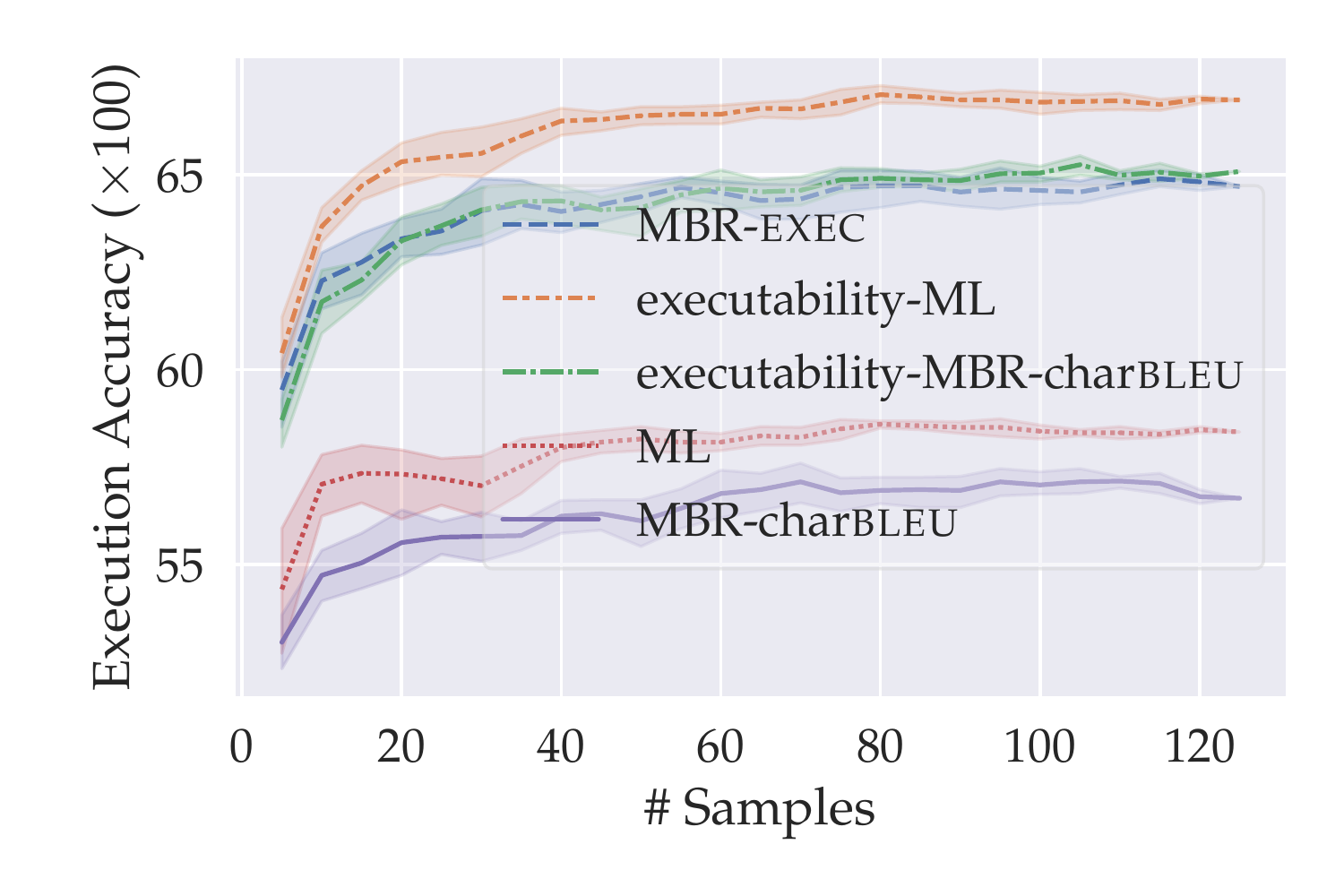}
        \caption{Spider}
    \end{subfigure}
    \caption{Comparison between applying methods to all possible candidates vs. applying methods to only executable candidates (best viewed in color), where executability-$X$ denotes applying selection criteria $X$ on executable candidates only. We did not include MBR-token\textsc{BLEU} and \maxavglikelihood and their combination with executability check in this figure for clarity -- full analysis on execution vs. executability can be found in appendix~\ref{appendix:full-executability}.}
    \label{fig:ablation-executability}
\end{figure}
\begin{figure}
    \centering
    \includegraphics[width=0.48\textwidth]{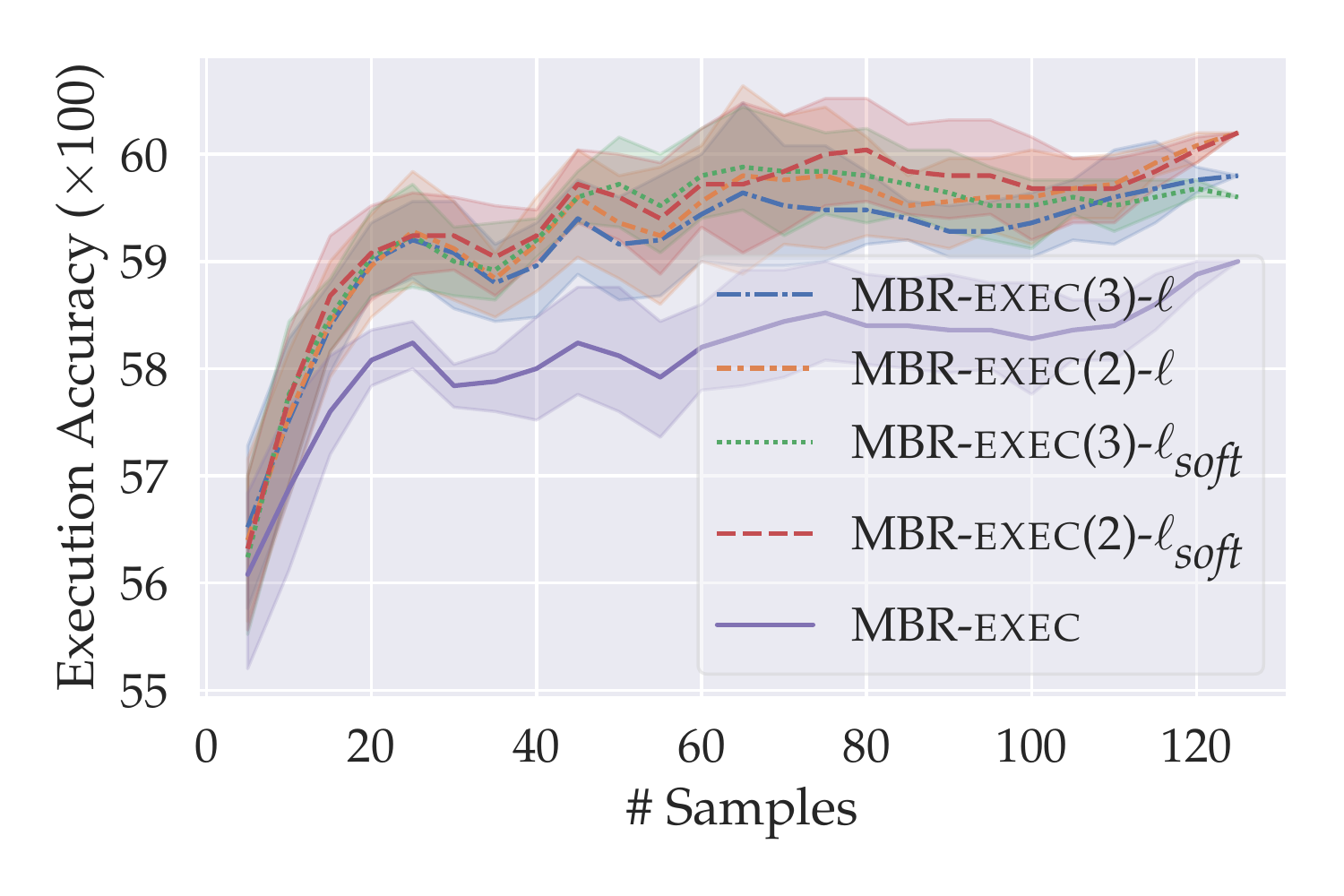}
    \caption{Execution accuracies with respect to sample size on the MBPP dataset, where the number in the parentheses denotes the number of test cases per problem used for \mbrexec. Best viewed in color.}
    \label{fig:mbpp-hard-soft-exec}
\end{figure}
\begin{figure*}[t]
    \centering
    \begin{subfigure}[t]{0.34\textwidth}
        \includegraphics[width=\textwidth]{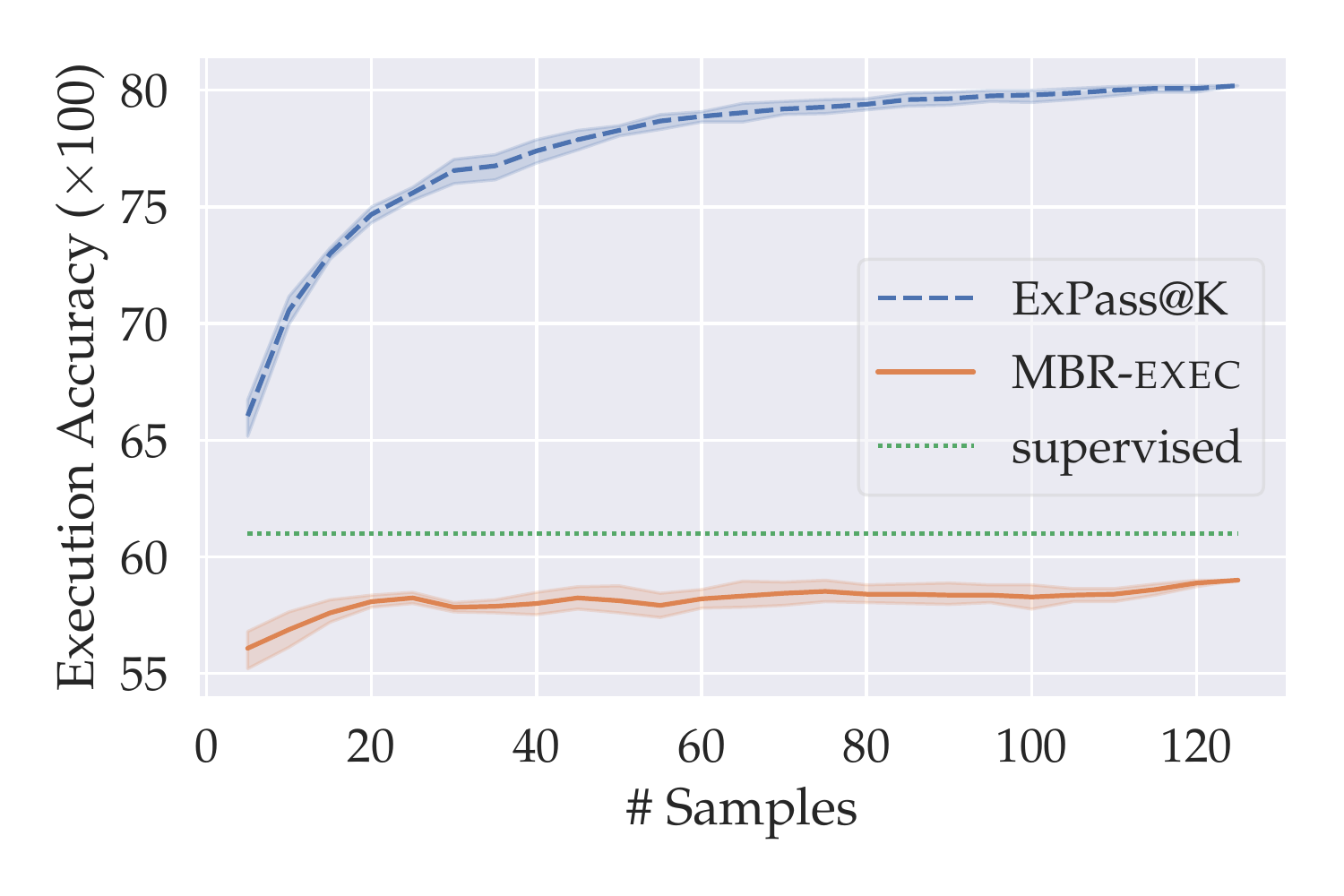}
        \caption{MBPP}
    \end{subfigure}
    \hspace{-10pt}
    \begin{subfigure}[t]{0.34\textwidth}
        \includegraphics[width=\textwidth]{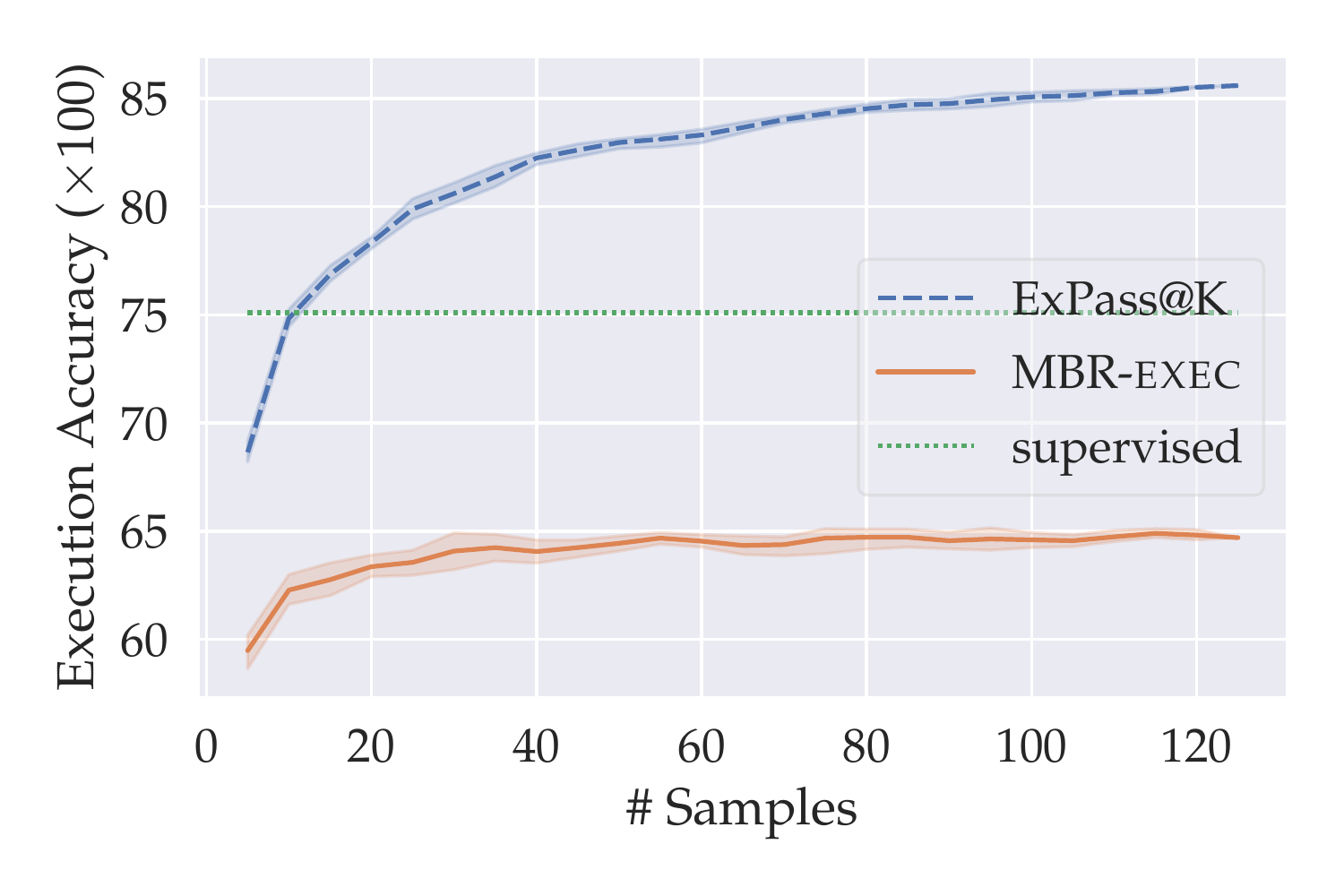}
        \caption{Spider}
    \end{subfigure}
    \hspace{-10pt}
    \begin{subfigure}[t]{0.34\textwidth}
        \includegraphics[width=\textwidth]{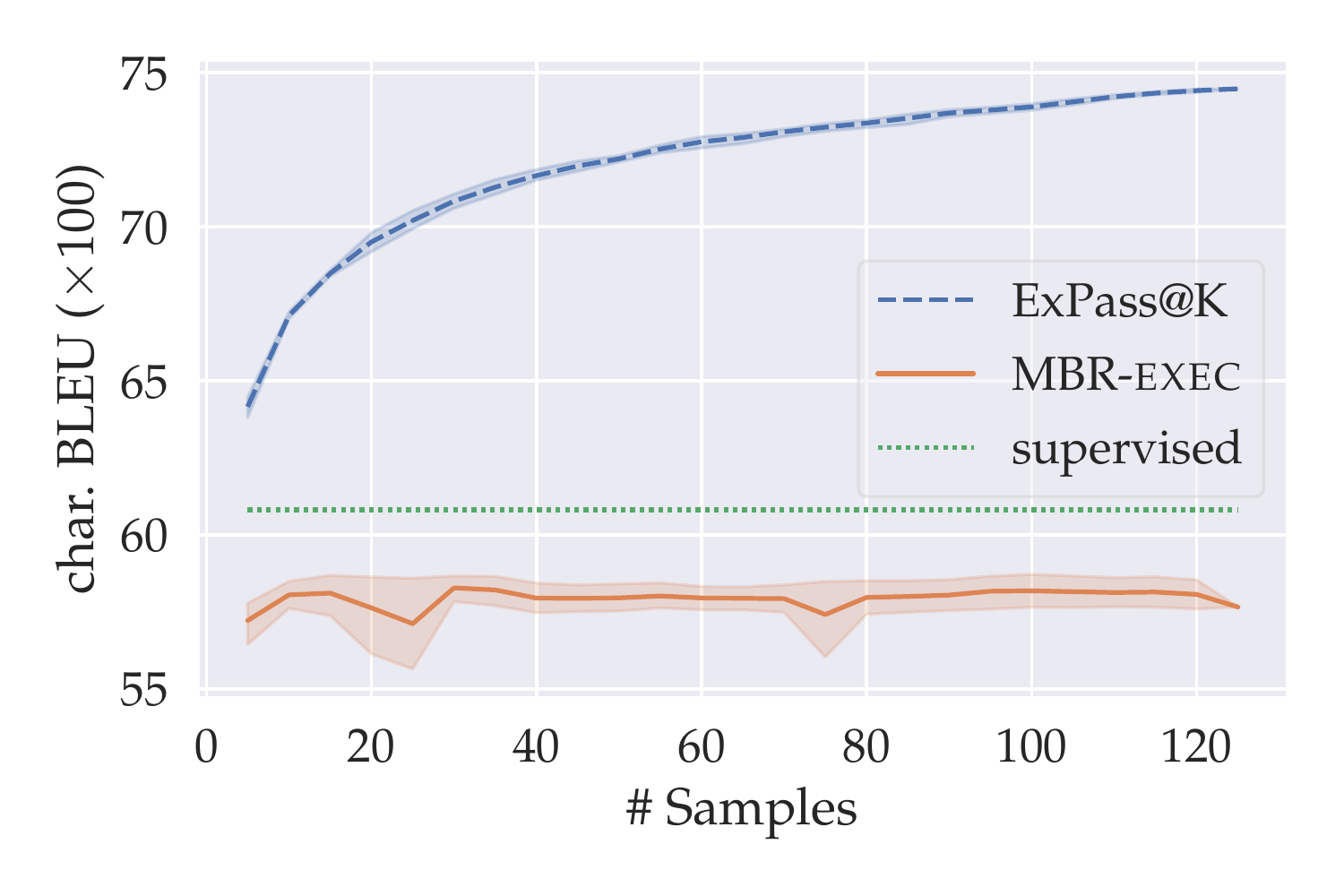}
        \caption{NL2Bash}
    \end{subfigure}
    \caption{Sample size--oracle performance curves on the considered datasets. We calculate each expected Pass@K with 5 different sets of candidates for each sample size, while using the same sets to perform \mbrexec for fair comparison. }
    \label{fig:oracle}
\end{figure*}
\subsection{Analysis}
\label{sec:expr-analysis}
We analyze the performance of \mbrexec from the following perspectives: the effectiveness across different sample collection temperatures (\S\ref{sec:expr-temperature}), the effectiveness of using groups of 3-shot prompts (\S\ref{sec:expr-15shot}) and the contribution of using execution results instead of simply checking the executability of programs (\S\ref{sec:expr-executability}).

\subsubsection{Effect of Sample Temperature}
\label{sec:expr-temperature}
We first compare sampling with temperature 0.3 to greedy decoding (i.e., temperature $\tau=0$) from the Codex model (Table~\ref{tab:0.3-vs-greedy}). When having the same number of examples, \mbrexec on sampled candidates with temperature 0.3 consistently reaches competitive or better performance than that on greedy decoded candidates. 

We plot the performance of \mbrexec for various sampling temperatures (Figure~\ref{fig:temperature}). Across datasets, we find that \mbrexec with a decoding temperature lower than 0.5 usually leads to reasonably good performance. When the temperature approaches 1.0, the results rapidly drop for all considered selection methods on MBPP and Spider; however, \maxavglikelihood generally achieves higher performance on NL2bash with a higher temperature. 

According to the evidences discussed above, we recommend to use sampling with a low temperature (specifically, lower than 0.5) for candidate sample collection, and perform \mbrexec for final program selection for better results. 

\subsubsection{Effect of Different 3-shot Prompts}
\label{sec:expr-15shot}
We analyze the necessity of choosing multiple groups of 3-shot instead of simply concatenating the available 15 examples as the prompt (Figure~\ref{fig:15-shot}).\footnote{We only include MBPP and NL2Bash results here as concatenating 15 Spider examples usually results in exceeding the token number limit of the pretrained models.} 
We allow different orders of the 15 examples when collecting samples. 
On both MBPP and NL2Bash datasets, we find that using different groups of 3-shot prompts clearly outperforms concatenating all 15 examples, suggesting that different groups of fewer-shot prompts followed by post-hoc decoding may be more effective than using all available examples for all time.

\subsubsection{Executability vs. Execution Results}
\label{sec:expr-executability}
We perform an ablation study to identify the contribution of execution results vs. program executability (Figure~\ref{fig:ablation-executability}) on the MBPP and Spider datasets.\footnote{We did not include NL2bash since \mbrexec does not really execute the commands. However, the comparison between \mbrexec and MBR-token\textsc{bleu} in Figure~\ref{fig:temperature}(c) shows that using an external bash parser as an executability estimator leads to more consistent and generally better performance.} We try to execute all candidates on the test cases, and perform baseline candidate methods only on the candidates that successfully execute within the time limit. On both datasets, we find that simply involving executability checking significantly helps improve the performance of all non-semantic feature--based selection methods; on Spider, applying \maxlikelihood over executable commands even outperforms \mbrexec across sample sizes. 

\subsubsection{Soft Loss as the Bayes Risk Function}
While all the above evaluations are based on executing one test case per problem, more test cases can lead to more accurate judgments of semantic equivalence between programs \citep{zhong-etal-2020-semantic}. 
Therefore, we introduce more test cases, and compare $\ell$ (\S\ref{sec:mbr-exec}) with $\ell_\textit{soft}$, a soft version of the loss function, as the Bayes risk function in \mbrexec. We define $\ell_\textit{soft}$ as follows:
\begin{align*}
    \ell_\textit{soft} (p_i, p_j) &= \frac{1}{|\mathcal{T}|}\sum_{t \in \mathcal{T}} \mathbbm{1}\left[p_i(t) \neq p_j(t)\right],
\end{align*}
which assesses equivalence based on the number of test cases that receive the same output. If there is only one test case available, $\ell$ and $\ell_\textit{soft}$ are equivalent.

We experiment with the MBPP dataset (Figure~\ref{fig:mbpp-hard-soft-exec}) as it provides three test cases per problem. While multiple test cases clearly outperforms \mbrexec with one test case across sample sizes, we did not find significant difference between $\ell_\textit{hard}$ and $\ell_\textit{soft}$, nor between using two or three test cases. 
\subsection{Oracle Performance}
\label{sec:expr-oracle}
We report the upper bound performance of all inference methods (Figure~\ref{fig:oracle}). Here, we define the expected Pass@K on one problem $q$ by 
\begin{align*}
    & \textit{ExPass@K}(q) \\
    = &\mathbb{E}_{|\mathcal{P}| = K} \left[ 
        \max_{p\in \mathcal{P}}
        \min_{t \in \mathcal{T}_q}
        \mathbbm{1}\left[
            p(t) = G(t)
        \right]
    \right],
\end{align*}
where $G(t)$ denotes the ground-truth output for test case input $t$. 
Intuitively, to calculate the performance upper bound, a problem $q$ is considered to be solved if there exists one program in the candidate sample set $P$ that passes all associated test cases $\mathcal{T}_q$. The dataset-level expected Pass@K is defined as the average expected Pass@K over all problems.

In addition, we report the supervised performance on these datasets, where all available training data are used for model training or finetuning: for MBPP, the results are from \citet{austin2021program}, where they use all 374 training examples to finetune their pretrained code model; for Spider, we compare to the current state-of-the-art result  \citep{scholak-etal-2021-picard}; for NL2Bash, we finetune GPT-2 \citep{radford2019language} with all training examples with the same prompting set up as Table~\ref{tab:prompt-format}.

However, it is worth noting that the upper bounds already outperform the state-of-the-art supervised performances on all datasets by a significant margin, when a reasonable amount of sample is given. This further demonstrates the effectiveness of the pretrained code models, and points out a potential next step in the direction: while such models are able to generate correct programs, designing effective inference algorithm may be a promising way towards translating natural language to code in real world applications.
\section{Discussion}
\label{sec:discussion}
We presented and systematically analyzed \mbrexec, an execution--based inference algorithm for pretrained language to code models, on datasets that cover three representative programming languages. Our results showed that doing execution, even with access only to inputs (not outputs) for test cases, or with only access to an executability checker, substantially helps improve the quality of generated programs especially in the settings that use execution accuracy as the evaluation metric (MBPP and Spider). Given the consistently strong performance, we suggest future work on program synthesis with large pretrained models consider \mbrexec as an effective selection algorithm. When we are not able to execute programs, or there are no test inputs available, our results suggest considering an alternative MBR metric (e.g., \mbrbleu) as the selection algorithm.

\section*{Limitations}
In this work, all selection methods are performed on top of a frozen pretrained code model \citep[Codex;][]{chen2021evaluating}. We note that incorporating execution information into the training or finetuning process of pretrained models may further help improve the performance. We leave the exploration of joint execution and training to future work.

\bibliography{anthology,custom}
\bibliographystyle{acl_natbib}

\appendix
\noindent {\Large \textbf{Appendices}}
\section{Example Prompts and Codex API Responses}
\label{appendix:prompts}
We include example 3-shot prompts and corresponding Codex responses that we used in our experiments, on the three datasets (Tables~\ref{tab:real-prompt-mbpp}, \ref{tab:real-prompt-spider}, \ref{tab:real-prompt-nl2bash}), where we format the prompts following the patterns presented in Table~\ref{tab:prompt-format}. Data shown in the tables are collected with the greedy decoding strategy (i.e., temperature = 0), and can be found in the first line of seed 0 in our released data for each test dataset. 
\begin{table*}[t]
    \centering \small
    \begin{tabular}{p{0.96\textwidth}}
    \toprule
    \textbf{MBPP: Prompt} \\
    \midrule ~\\[-22pt]
    \begin{lstlisting}
<info>assert camel_to_snake('GoogleAssistant') == 'google_assistant'</info>
<text>Write a function to convert camel case string to snake case string by using regex.</text>
<code>import re
def camel_to_snake(text):
  str1 = re.sub('(.)([A-Z][a-z]+)', r'\\1_\\2', text)
  return re.sub('([a-z0-9])([A-Z])', r'\\1_\\2', str1).lower()</code>
<info>assert sort_dict_item({(5, 6) : 3, (2, 3) : 9, (8, 4): 10, (6, 4): 12} ) == {(2, 3): 9, (6, 4): 12, (5, 6): 3, (8, 4): 10}</info>
<text>Write a function to sort dictionary items by tuple product of keys for the given dictionary with tuple keys.</text>
<code>def sort_dict_item(test_dict):
  res = {key: test_dict[key] for key in sorted(test_dict.keys(), key = lambda ele: ele[1] * ele[0])}
  return  (res) 
</code>
<info>assert reverse_list_lists([[1, 2, 3, 4], [5, 6, 7, 8], [9, 10, 11, 12], [13, 14, 15, 16]])==[[4, 3, 2, 1], [8, 7, 6, 5], [12, 11, 10, 9], [16, 15, 14, 13]]</info>
<text>Write a function to reverse each list in a given list of lists.</text>
<code>def reverse_list_lists(lists):
    for l in lists:
        l.sort(reverse = True)
    return lists </code>
<info>assert remove_Occ(\"hello\",\"l\") == \"heo\"</info>
<text>Write a python function to remove first and last occurrence of a given character from the string.</text>
<code>
    \end{lstlisting} \\[-15pt]
    \midrule 
    \midrule
    \textbf{MBPP: Response} \\
    \midrule ~\\[-22pt]
    \begin{lstlisting}
def remove_Occ(str1,ch):
    return str1[:str1.index(ch)] + str1[str1.rindex(ch)+1:]</code>
    \end{lstlisting} \\[-15pt]
    \bottomrule
    \end{tabular}
    \caption{\label{tab:real-prompt-mbpp} MBPP example prompt and response from Codex: we use the first assertion in the dataset as the extra information (i.e., \texttt{[INFO]} in Table~\ref{tab:prompt-format}). The content in the last \texttt{<info>...</info>} and \texttt{<text>...</text>} marks in the prompt corresponds to the test problem.}
\end{table*}

\begin{table*}[t]
    \centering \small
    \begin{tabular}{p{0.96\textwidth}}
    \toprule
    \textbf{Spider: Prompt} \\
    \midrule ~\\[-22pt]
    \begin{lstlisting}
<info>e_learning | * | Course_Authors_and_Tutors : author_id , author_tutor_ATB , login_name , password , personal_name , middle_name , family_name , gender_mf , address_line_1 | Students : student_id , date_of_registration , date_of_latest_logon , login_name , password , personal_name , middle_name , family_name | Subjects : subject_id , subject_name | Courses : course_id , author_id , subject_id , course_name , course_description | Student_Course_Enrolment : registration_id , student_id , course_id , date_of_enrolment , date_of_completion | Student_Tests_Taken : registration_id , date_test_taken , test_result</info>
<text>Which course authors teach two or more courses? Give me their addresses and author IDs.</text>
<code>SELECT T1.address_line_1 ,  T2.author_id FROM Course_Authors_and_Tutors AS T1 JOIN Courses AS T2 ON T1.author_id  =  T2.author_id GROUP BY T2.author_id HAVING Count(*)  >=  2</code>
<info>flight_1 | * | flight : flno , origin , destination , distance , departure_date , arrival_date , price , aid | aircraft : aid , name , distance | employee : eid , name , salary | certificate : eid , aid</info>
<text>Show origin and destination for flights with price higher than 300.</text>
<code>SELECT origin ,  destination FROM Flight WHERE price  >  300</code>
<info>driving_school | * | Addresses : address_id , line_1_number_building , city , zip_postcode , state_province_county , country | Staff : staff_id , staff_address_id , nickname , first_name , middle_name , last_name , date_of_birth , date_joined_staff , date_left_staff | Vehicles : vehicle_id , vehicle_details | Customers : customer_id , customer_address_id , customer_status_code , date_became_customer , date_of_birth , first_name , last_name , amount_outstanding , email_address , phone_number , cell_mobile_phone_number | Customer_Payments : customer_id , datetime_payment , payment_method_code , amount_payment | Lessons : lesson_id , customer_id , lesson_status_code , staff_id , vehicle_id , lesson_date , lesson_time , price</info>
<text>When did the staff member with first name as Janessa and last name as Sawayn leave the company?</text>
<code>SELECT date_left_staff FROM Staff WHERE first_name = \"Janessa\" AND last_name = \"Sawayn\";</code>
<info>concert_singer | * | stadium : Stadium_ID , Location , Name , Capacity , Highest , Lowest , Average | singer : Singer_ID , Name , Country , Song_Name , Song_release_year , Age , Is_male | concert : concert_ID , concert_Name , Theme , Stadium_ID , Year | singer_in_concert : concert_ID , Singer_ID</info>
<text>How many singers do we have?</text>
<code>
    \end{lstlisting} \\[-15pt]
    \midrule 
    \midrule
    \textbf{Spider: Response} \\
    \midrule ~\\[-22pt]
    \begin{lstlisting}
SELECT COUNT(*) FROM singer;</code>
    \end{lstlisting} \\[-15pt]
    \bottomrule
    \end{tabular}
    \caption{\label{tab:real-prompt-spider} Spider example prompt and response from Codex: following \citet{xie2022unifiedskg}, we use the concatenation of corresponding table and column names in the dataset as the extra information (i.e., \texttt{[INFO]} in Table~\ref{tab:prompt-format}).. The content in the last \texttt{<info>...</info>} and \texttt{<text>...</text>} marks in the prompt corresponds to the test problem.}
\end{table*}

\begin{table*}[t]
    \centering \small
    \begin{tabular}{p{0.96\textwidth}}
    \toprule
    \textbf{NL2Bash: Prompt} \\
    \midrule ~\\[-22pt]
    \begin{lstlisting}
<text>Print file information of command \"bash\"</text>
<code>echo $(ls -l $(which bash))</code>
<text>Recursively change the owner and group of all files in \"/your/directory/to/fuel/\" to \"nginx\"</text>
<code>chown nginx:nginx /your/directory/to/fuel/ -R</code>
<text>Copy \"src/prog.js\" and \"images/icon.jpg\" to \"/tmp/package\" keeping relative path names</text>
<code>rsync -R src/prog.js images/icon.jpg /tmp/package</code>
<text>Adds execution permissions on a script ./etc/bash_completion within Homebrew home folder path.</text>
<code>
    \end{lstlisting} \\[-15pt]
    \midrule 
    \midrule
    \textbf{NL2Bash: Response} \\
    \midrule ~\\[-22pt]
    \begin{lstlisting}
chmod +x /usr/local/etc/bash_completion</code>
    \end{lstlisting} \\[-15pt]
    \bottomrule
    \end{tabular}
    \caption{\label{tab:real-prompt-nl2bash} NL2Bash example prompt and response from Codex: we did not use any extra information. The content in the last \texttt{<text>...</text>} marks in the prompt corresponds to the test problem.}
\end{table*}

\section{Full Analysis on Executability vs. Execution Result}
\label{appendix:full-executability}
\begin{figure}[t]
    \centering
    \begin{subfigure}[t]{0.45\textwidth}
        \includegraphics[width=\textwidth]{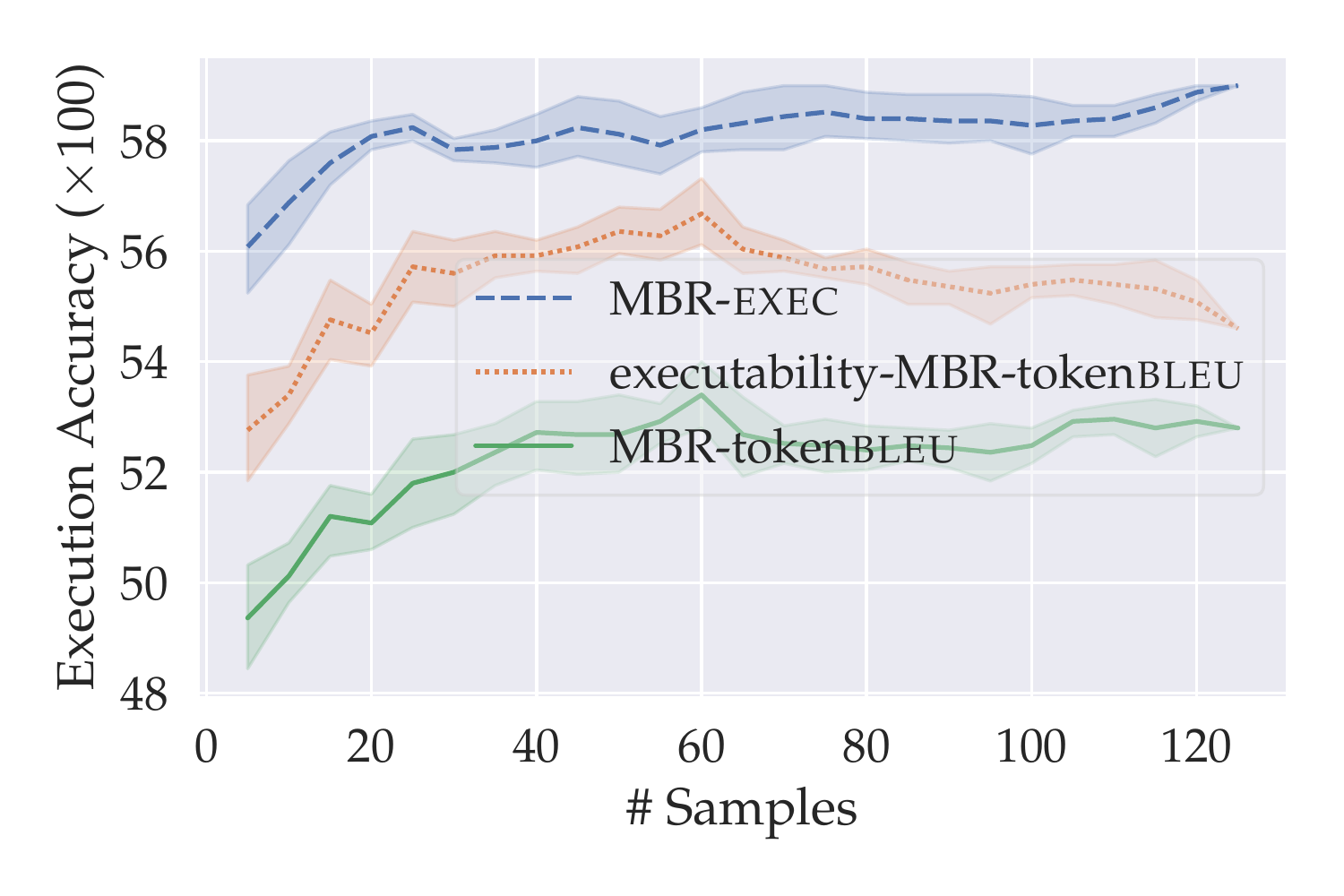}
        \caption{MBPP (MBR-token\textsc{BLEU})}
    \end{subfigure}
    \hspace{-10pt}
    \begin{subfigure}[t]{0.45\textwidth}
        \includegraphics[width=\textwidth]{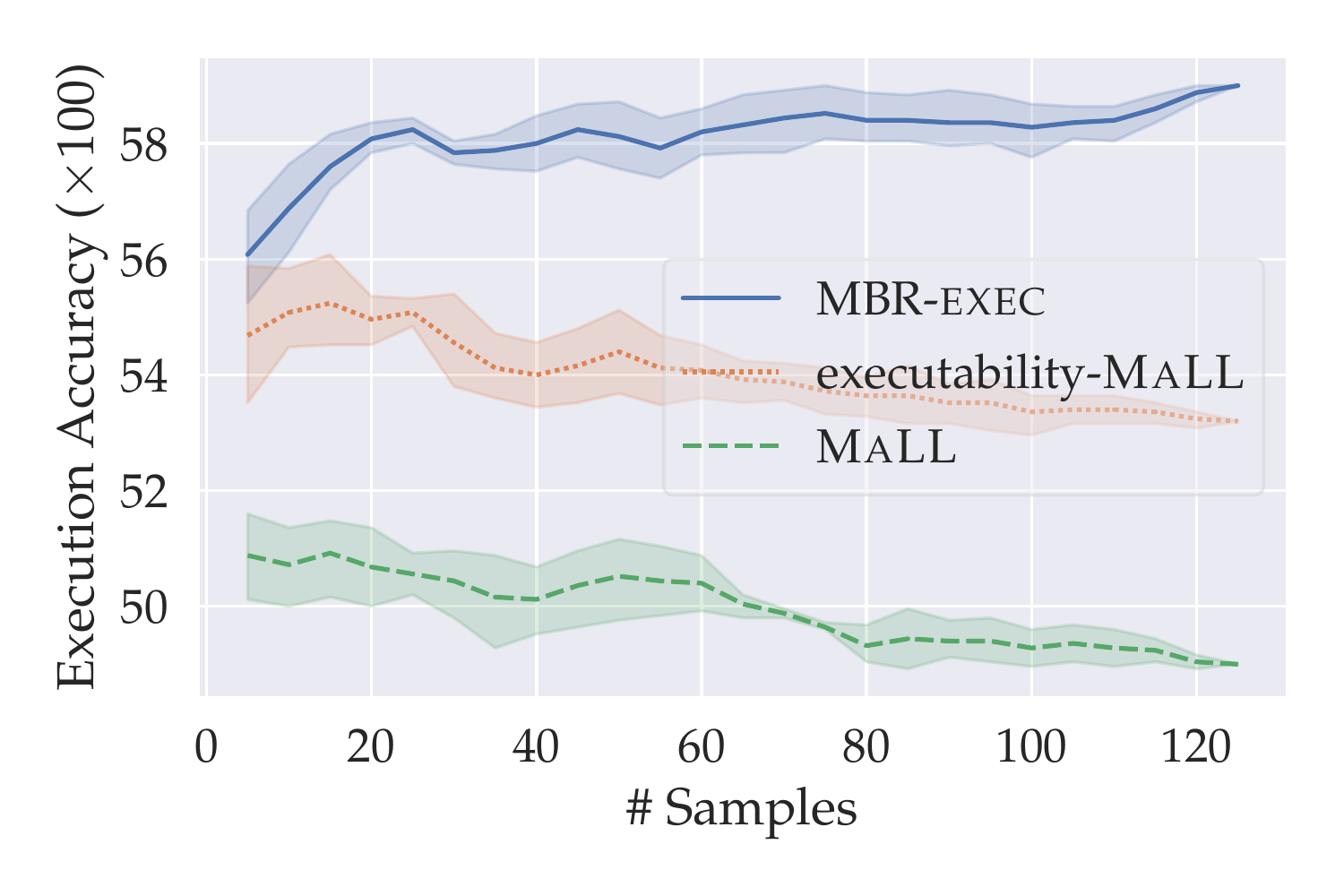}
        \caption{MBPP (\maxavglikelihood)}
    \end{subfigure}
    \hspace{-10pt}
    \begin{subfigure}[t]{0.45\textwidth}
        \includegraphics[width=\textwidth]{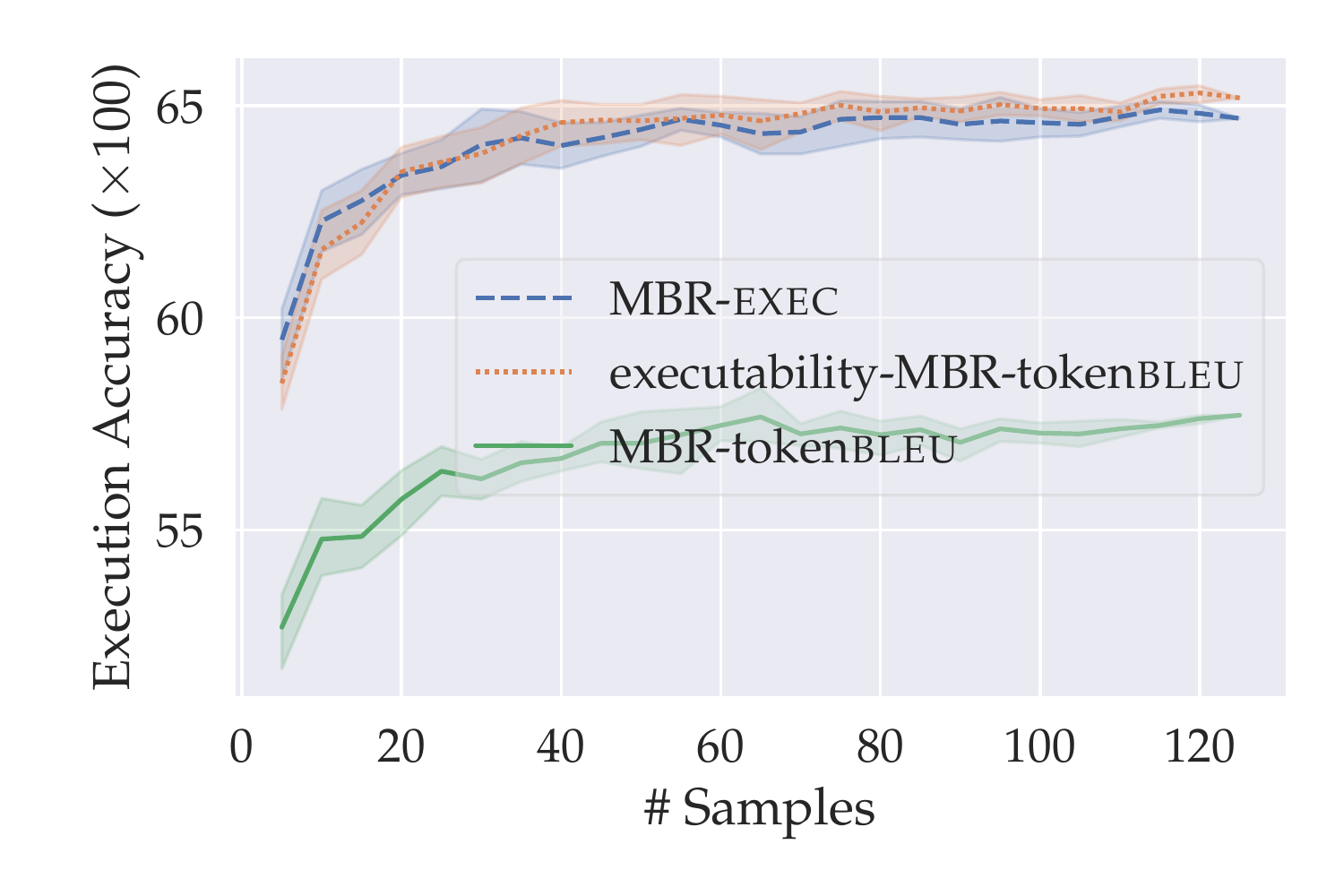}
        \caption{Spider (MBR-token\textsc{BLEU})}
    \end{subfigure}
    \hspace{-10pt}
    \begin{subfigure}[t]{0.45\textwidth}
        \includegraphics[width=\textwidth]{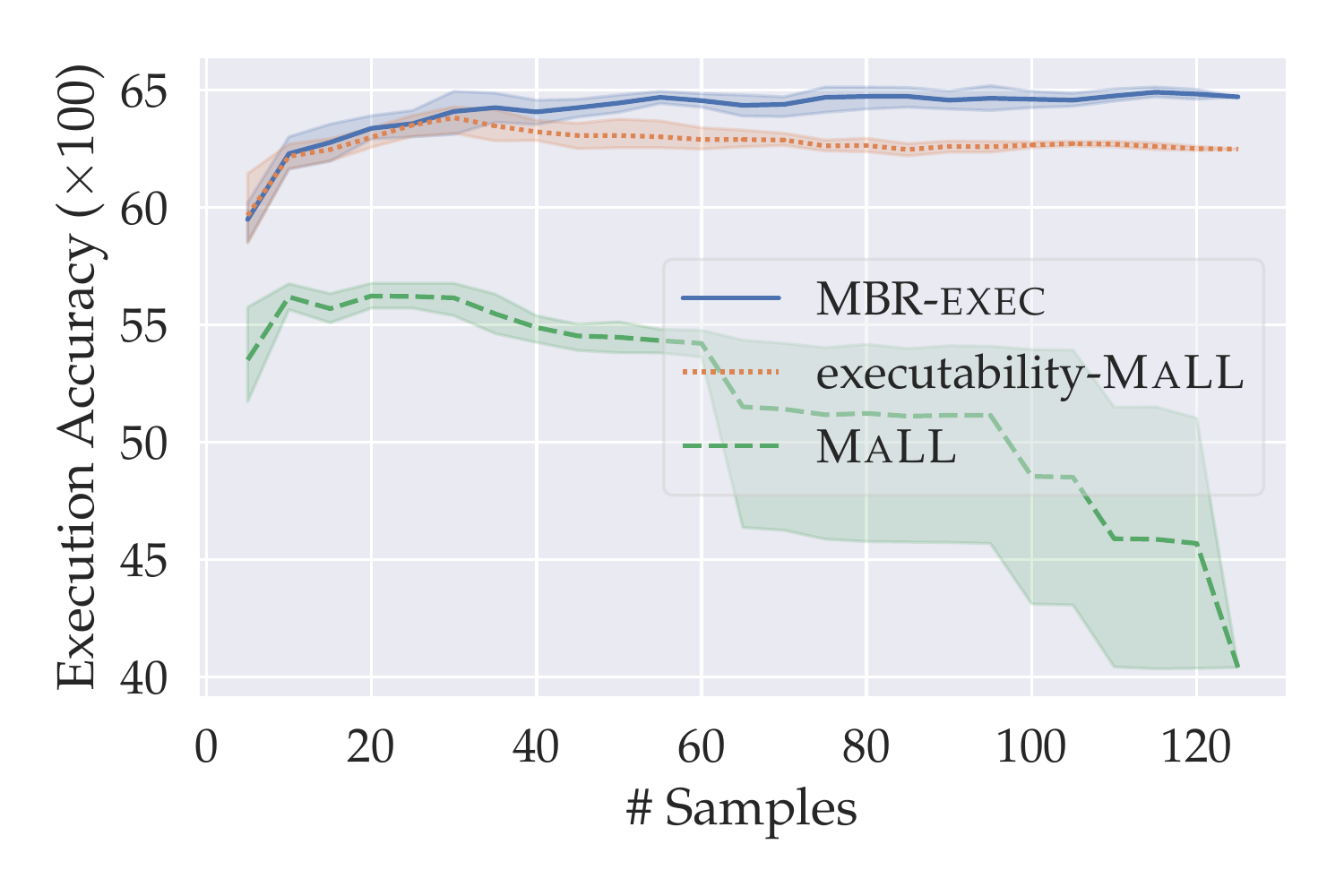}
        \caption{Spider (\maxavglikelihood)}
    \end{subfigure}
    \caption{Comparison between applying methods to all possible candidates vs. applying methods to only executable candidates (best viewed in color), where executability-$X$ denotes applying selection criteria $X$ on executable candidates only. We also include the curves of \mbrexec for comparison.}
    \label{fig:ablation-executability-full}
\end{figure}
We report the comparison between MBR-token\textsc{BLEU} and \maxavglikelihood vs. their combination with executability check (Figure~\ref{fig:ablation-executability-full}; in complementary to Figure~\ref{fig:ablation-executability}), where we observe that an executability checker is an effective filter to improve execution accuracies for both datasets (MBPP and Spider).

\end{document}